\documentclass[10pt]{article} % For LaTeX2e

\pdfoutput=1
\PassOptionsToPackage{compress, sort}{natbib}

\usepackage[accepted]{tmlr}
\usepackage[utf8]{inputenc} % allow utf-8 input
\usepackage[T1]{fontenc}    % use 8-bit T1 fonts
\usepackage{hyperref}       % hyperlinks
\usepackage{enumitem}
\usepackage{url}            % simple URL typesetting
\usepackage{xspace}
\usepackage{booktabs}       % professional-quality tables
\usepackage{amsfonts}       % blackboard math symbols
\usepackage{nicefrac}       % compact symbols for 1/2, etc.
\usepackage{microtype}      % microtypography
\usepackage[noend]{algpseudocode}
\usepackage[ruled,linesnumbered,vlined]{algorithm2e}
\usepackage{multirow}
\usepackage{multicol}
\usepackage{amsmath}
\usepackage{amssymb}
\usepackage{amsthm}
\usepackage{graphicx}
\usepackage{subcaption}
\usepackage{wrapfig}
\usepackage{mathtools}
\usepackage[capitalise, noabbrev]{cleveref}
\usepackage{tikz}
\hypersetup{
    colorlinks = true,
    citecolor = [RGB]{0,130,130},
    urlcolor = [RGB]{200,0,100}
}

\makeatletter \renewcommand\paragraph{\@startsection{paragraph}{4}{\z@} {2mm \@plus1ex \@minus.2ex} {-0.7em} {\normalfont\normalsize\bfseries}} \makeatother

\newcommand\overstar[1]{\ThisStyle{\ensurestackMath{%
  \setbox0=\hbox{$\SavedStyle#1$}%
  \stackengine{0pt}{\copy0}{\kern.2\ht0\smash{\SavedStyle*}}{O}{c}{F}{T}{S}}}}

\newtheorem{theorem}{Theorem}
\newtheorem{mydefinition}[theorem]{Definition}
\newtheorem{loosedefinition}[theorem]{(Loose) Definition}

\newcommand{\STAB}[1]{\begin{tabular}{@{}c@{}}#1\end{tabular}}

\newcommand{\dataset}[0]{\ensuremath{\mathcal{D}}\xspace}
\newcommand{\distill}[0]{\ensuremath{\mathcal{D}_{\mathsf{syn}}}\xspace}

\newcommand{\ie}[0]{\emph{i.e.}\xspace}
\newcommand{\etc}[0]{\emph{etc.}\xspace}
\newcommand{\eg}[0]{\emph{e.g.}\xspace}
\newcommand{\vs}[0]{\emph{vs.}\xspace}
\newcommand{\wrt}[0]{\emph{w.r.t.}\xspace}
\newcommand{\aka}[0]{\emph{a.k.a.}\xspace}

\newcommand{\expectation}[2]{\underset{#1}{\mathbb{E}} \left[#2\right]}

\title{Data Distillation: A Survey}
\author{\name Noveen Sachdeva \email nosachde@ucsd.edu \\
      \addr Computer Science \& Engineering\\
      University of California, San Diego
      \AND
      \name Julian McAuley \email jmcauley@ucsd.edu \\
      \addr Computer Science \& Engineering\\
      University of California, San Diego%
      }

\def\month{07}  % Insert correct month for camera-ready version
\def\year{2023} % Insert correct year for camera-ready version
\def\openreview{\url{https://openreview.net/forum?id=lmXMXP74TO}} % Insert correct link to OpenReview for camera-ready version

\begin{document}

\maketitle

\begin{abstract}
    The popularity of deep learning has led to the curation of a vast number of massive and multifarious datasets. Despite having close-to-human performance on individual tasks, training parameter-hungry models on large datasets poses multi-faceted problems such as (a) high model-training time; (b) slow research iteration; and (c) poor eco-sustainability. As an alternative, \emph{data distillation} approaches aim to synthesize terse data summaries, which can serve as effective drop-in replacements of the original dataset for scenarios like model training, inference, architecture search, \etc
    % What we do in this survey -- extend more?
    In this survey, we present a formal framework for data distillation, along with providing a detailed taxonomy of existing approaches. Additionally, we cover data distillation approaches for different data modalities, namely images, graphs, and user-item interactions (recommender systems), while also identifying current challenges and future research directions. % \looseness=-1
\end{abstract}

\section{Introduction}
\begin{loosedefinition} \label{def:data_distillation}
    {\normalfont \textbf{(Data distillation)}} Approaches that aim to synthesize tiny and high-fidelity data summaries which distill the most important knowledge from a given target dataset. Such distilled summaries are optimized to serve as effective drop-in replacements of the original dataset for efficient and accurate data-usage applications like model training, inference, architecture search, etc.
\end{loosedefinition}
\vspace{0.1cm}

% Reference figure 1
The recent ``scale-is-everything'' viewpoint \citep{scaling_1, scaling_2, scaling_3}, argues that training bigger models (\ie, consisting of a higher number of parameters) on bigger datasets, and using larger computational resources is the sole key for advancing the frontier of artificial intelligence. Such studies observe and hypothesize the generalizability of neural networks as a power-law \wrt the aforementioned factors, albeit with small exponents.
On the other hand, a reasonable argument is that a principled and well-reasoned solution will be more amenable to various scaling-laws, thereby leading to faster progress. 
Data distillation (Definition \ref{def:data_distillation}) is clearly a task rooted in the latter school of thought by introducing the \emph{fidelity of data} as in important covariate in such neural scaling-laws. 
\citet{data_quality} demonstrate this viewpoint analytically by using simple heuristics to
% where they largely improve neural scaling laws by 
prune away data with low measures of signal for model training.
Clearly, the scale viewpoint still holds, in that if we keep increasing the amount of data (albeit now compressed and of higher quality), we will observe an improvement in both upstream and downstream generalization, but at a faster rate.

% re-usability of knowledge
\paragraph{Motivation.} A terse, high-quality data summary has use cases from a variety of standpoints. First and foremost, it leads to a faster model-training procedure. In turn, faster model training equates to (1) compute-cost saving and expedited research iterations, \ie, the investigative procedure of manually experimenting different ideas; 
% (2) compute-cost saving, \ie, reductions in model-training time also leads to reductions in the amount- and time-of compute resources used; 
and (2) improved eco-sustainability, \ie, lowering the amount of compute time directly leads to a lower carbon footprint from running power-hungry accelerated hardware \citep{chasing_carbon}. Additionally, a small data summary democratizes the entire pipeline, as more people can train state-of-the-art algorithms on reasonably accessible hardware using the data summary. Finally, 
a high-quality data summary indirectly also accelerates orthogonal procedures like neural architecture search \citep{darts_nas}, approximate nearest neighbour search \citep{ann}, knowledge distillation \citep{knowledge_distillation}, \etc, where the procedure needs to iterate over the entire dataset multiple times.
% a high-quality data summary also inherently promotes faster knowledge transfer. More specifically, all learning algorithms rely on a collected source of data to estimate their parameters from. Having a high-quality, reusable data summary can be used to train a wide variety of learning algorithms for individual downstream use-cases, without being constrained to share knowledge amongst each other. 

\begin{figure*}[t!] \centering
    \centering
    \includegraphics[width=0.7\linewidth]{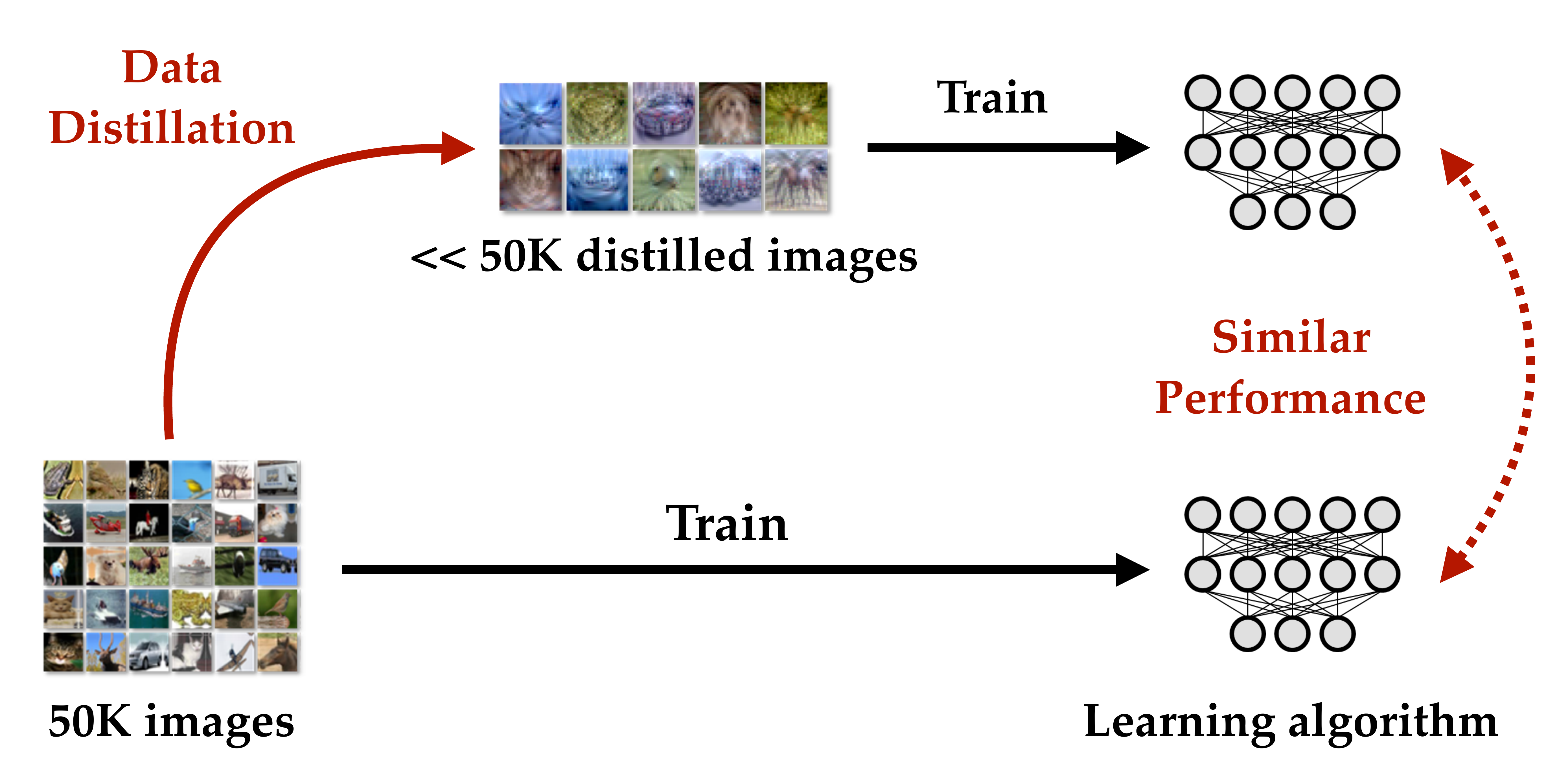}
    % \vspace{4pt}
    \renewcommand\figurename{\href{https://www.noveens.com/images/dd_survey/overview.pdf}{[HQ Image Link]} Figure}
    \caption{The premise of data distillation demonstrated using an image dataset.}
    \vspace{-0.1cm}
    \label{fig:overview}
\end{figure*} 

\paragraph{Comparison with data pruning.} Another reasonable avenue for summarizing large datasets is \emph{pruning} away \emph{low-quality} data which presumably does not carry large amount of \emph{signal} to be captured during model-training. 
% in-depth
The primary challenge for such data pruning approaches (\aka coreset construction) lies in tagging the \emph{hardness} of each data-point which can be used for subsequent pruning (typically in a greedy fashion). Prominent data-pruning approaches propose heuristics for the same, relying on concepts such as shapley values \citep{data_shapley}, confidence scores \citep{svp}, error-contribution \citep{forgetting}, feature-space geometry \citep{data_quality,herding,semdedup}, \etc
% Coresets
Another line of work builds on the advances in submodular optimization (see \citet{bilmes_submodularity} for a review) to approximately solve the NP-Hard combinatorial optimization of selecting the subset that maximizes a set-level \emph{goodness} function, when such goodness functions are provably submodular \citep{craig,grad_match,selcon_coreset}.
% comparison to DD
Notably, such data pruning methodologies inherently share the same goal as data distillation but are severely restricted due to only retaining data already in the target dataset, thereby leading to finite expressivity and hence, generally, lower sample-fidelity (see \citet{data_pruning_limitations} for a deeper theoretical outlook on the fundamental limitations of data pruning). Further, recent empirical studies of data pruning methodologies \citep{deepcore} demonstrate that the efficacy of such data pruning heuristics rarely and irregularly transfers to practical scenarios, with random downsampling being a hard baseline.

\paragraph{Comparison with knowledge distillation \& transfer learning.} Despite inherently distilling some notion of \emph{knowledge}, we would like to highlight both \emph{knowledge distillation} and \emph{transfer learning} are orthogonal procedures to data distillation, and can potentially work in conjunction to perform both tasks more efficiently. More specifically, knowledge distillation \citep{knowledge_distillation} entails distilling the knowledge from a trained teacher network into a smaller student network. On the other hand, transfer learning \citep{transfer_learning} focuses on transferring knowledge across similar tasks, \eg, from image classification to image segmentation. Orthogonally, data distillation aims to distill the knowledge from a given dataset into a terse data summary. Such data summaries can be used \emph{in conjunction} with knowledge distillation or transfer learning procedures for both (1) faster learning of the teacher models; and (2) faster knowledge transfer to the student models. The same holds true for model compression techniques \citep{model_compression}, where similar to knowledge distillation, the goal is to reduce model storage size rather than reducing the training time or increasing the sample-fidelity.

\begin{figure*}[t!] \centering
    \begin{tikzpicture}[font=\fontfamily{ppl}]
    \node [anchor=west] (meta) at (0.48,-0.3) {\normalfont \textbf{\cref{sec:meta_model}}};
    \node [anchor=west] (grad) at (3.88,-0.3) {\normalfont \textbf{\cref{sec:grad_matching}}};
    \node [anchor=west] (traj) at (7.2,-0.3) {\normalfont \textbf{\cref{sec:traj_matching}}};
    \node [anchor=west] (dist) at (10.55,-0.3) {\normalfont \textbf{\cref{sec:distr_matching}}};
    \begin{scope}% [xshift=1.5cm]
        \node[anchor=south west,inner sep=0] (image) at (0,0) {\includegraphics[width=0.8\textwidth]{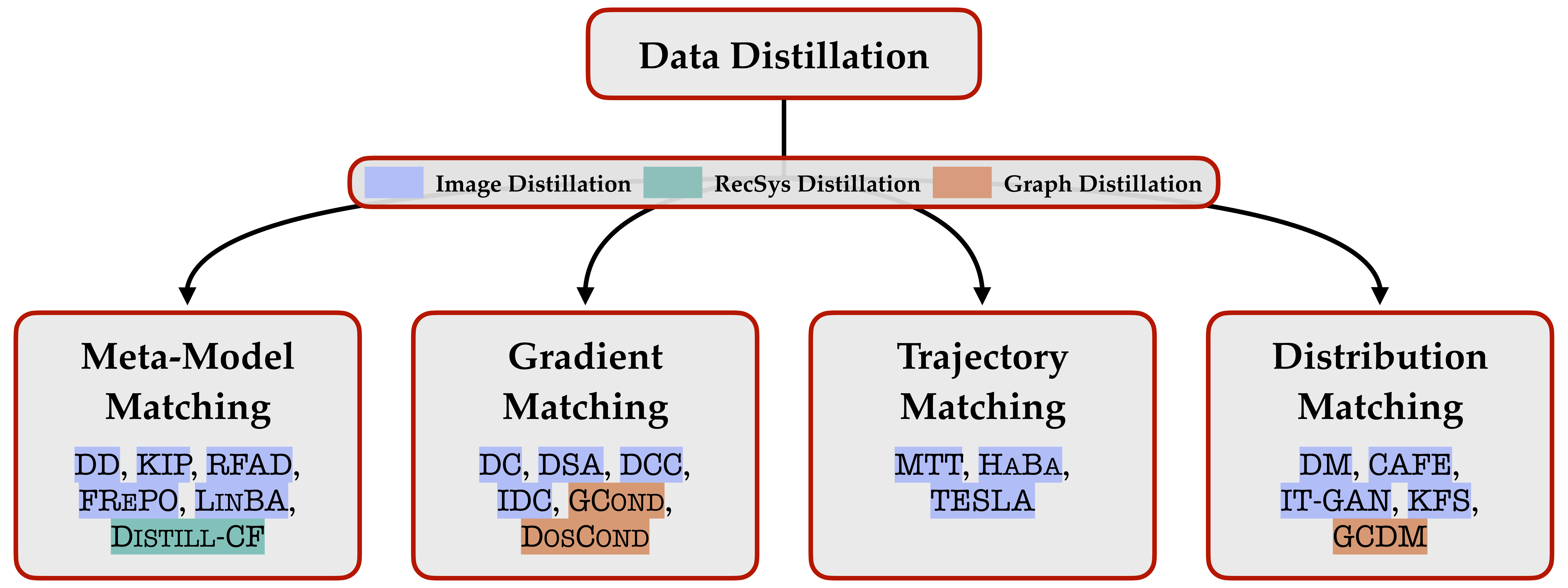}};
    \end{scope}
    \end{tikzpicture}%
    % \vspace{4pt}
    \renewcommand\figurename{\href{https://www.noveens.com/images/dd_survey/taxonomy.pdf}{[HQ Image Link]} Figure}
    \caption{A taxonomy of existing data distillation approaches.}
    \label{fig:taxonomy}
    \vspace{-4pt}
\end{figure*}
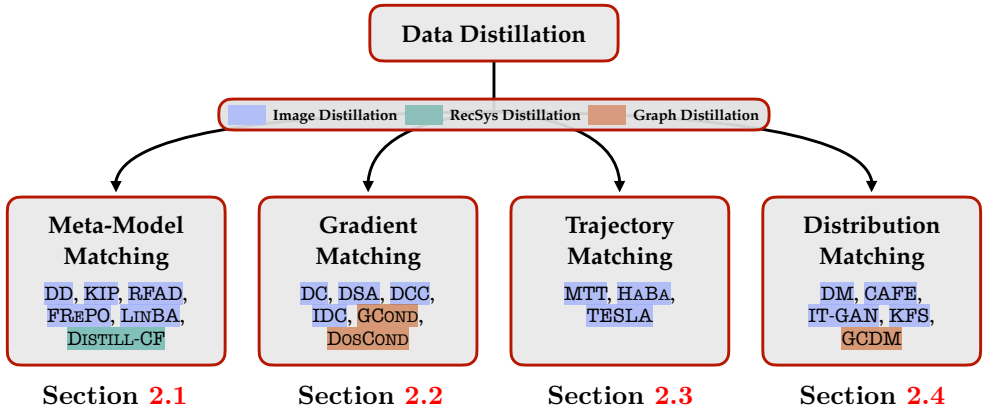

\vspace{0.001cm}
\emph{In this survey}, we intend to provide a succinct overview of various data distillation frameworks across different data modalities. We start by presenting a formal data distillation framework
% , along with discussing commonly employed optimization techniques 
in \cref{sec:framework}, and present technicalities of various existing techniques. We classify all data distillation techniques into four categories (see \cref{fig:taxonomy} for a taxonomy) and provide a detailed empirical comparison of image distillation techniques in \cref{tab:main_results}. Subsequently, in \cref{sec:data_modalities}, we discuss existing data distillation frameworks for synthesizing data of different modalities, as well as outlining the associated challenges. In \cref{sec:applications}, we discuss alternative applications of synthesizing a high-fidelity data summary rather than simply accelerating model training along with pointers to existing work. Finally, in \cref{sec:challenges_future_work}, we conclude by presenting common pitfalls in existing data distillation techniques, along with proposing interesting directions for future work.

\section{The Data Distillation Framework} \label{sec:framework}

Before going into the specifics of data distillation, we start by outlining useful notation. Let $\dataset \triangleq \{ (x_i, y_i) \}_{i=1}^{|\dataset|}$ be a given dataset which needs to be distilled, where $x_i \in \mathcal{X}$ are the set of input features, and $y_i \in \mathcal{Y}$ is the desired label for $x_i$. For classification tasks, let $\mathcal{C}$ be the set of unique classes in $\mathcal{Y}$, and $\dataset^c \triangleq \{ (x_i, y_i) ~|~ y_i = c\}_{i=1}^{|\dataset|}$ be the subset of $\dataset$ with class $c$. We also define the matrices $\mathbf{X} \triangleq [x_i]_{i=1}^{|\dataset|}$ and $\mathbf{Y} \triangleq [y_i]_{i=1}^{|\dataset|}$ for convenience. Given a data budget $n \in \mathbb{Z}^+$, data distillation techniques aim to synthesize a high-fidelity data summary $\distill \triangleq \{ (\Tilde{x}_i, \Tilde{y}_i) \}_{i=1}^{n}$ such that $n \ll |\dataset|$. We define $\distill^c$, $\mathbf{X}_{\mathsf{syn}}$, and $\mathbf{Y}_{\mathsf{syn}}$ similarly as defined for \dataset. Let $\Phi_{\theta} : \mathcal{X} \mapsto \mathcal{Y}$ represent a learning algorithm parameterized by $\theta$. We also assume access to a twice-differentiable cost function $l : \mathcal{Y} \times \mathcal{Y} \mapsto \mathbb{R}$, and define $\mathcal{L}_{\dataset}(\theta) \triangleq \mathbb{E}_{(x, y) \sim \dataset}[l(\Phi_{\theta}(x), y)]$ for convenience. Notation is also summarized in \cref{appendix:details}. Notably, since \dataset and \distill share the same data domain ($\mathcal{X}$), under reasonable systems' assumptions, training $\Phi$ using gradient descent (GD) on \distill will have a $\frac{|\dataset|}{n} \times$ training-time speedup compared to training $\Phi$ on \dataset. 

For the sake of uniformity, we refer to the data synthesized by data distillation techniques as a \emph{data summary} henceforth. Inspired by the definition of coresets \citep{coreset_general}, we formally define an \ $\epsilon-$approximate data summary, and the data distillation task as follows:

\begin{mydefinition} \label{def:data_quality}
    {\normalfont \textbf{(\boldmath{$\epsilon-$}approximate data summary)}} Given a learning algorithm $\Phi$, let $\theta^{\dataset}$, $\theta^{\distill}$ represent the optimal set of parameters for $\Phi$ estimated on $\dataset$ and $\distill$, and $\epsilon \in \mathbb{R}^+$; we define an $\epsilon-$approximate data summary as one which satisfies:
    \begin{equation} \label{eqn:eps_approx_summary}
        \operatorname{sup} ~ \left\{ ~ \left\vert \ l\left(\Phi_{\theta^{\dataset}}(x), y\right) - l\left(\Phi_{\theta^{\distill}}(x), y\right) \ \right\vert ~ \right\}_{\raisebox{6pt}{$\substack{x \sim \mathcal{X}\\y \sim \mathcal{Y}}$}} ~ \leq ~ \epsilon
    \end{equation}
\end{mydefinition}

\begin{mydefinition} \label{def:data_distillation_technical}
    {\normalfont \textbf{(Data distillation)}} Given a learning algorithm $\Phi$, let $\theta^{\dataset}$, $\theta^{\distill}$ represent the optimal set of parameters for $\Phi$ estimated on $\dataset$ and $\distill$; we define data distillation as optimizing the following:
    \begin{equation} \label{eqn:dd_optimization}
        \underset{\distill, n}{\operatorname{arg} \operatorname{min}} \left( \operatorname{sup} ~ \left\{ ~ \left\vert \ l\left(\Phi_{\theta^{\dataset}}(x), y\right) - l\left(\Phi_{\theta^{\distill}}(x), y\right) \ \right\vert ~ \right\}_{\raisebox{6pt}{$\substack{x \sim \mathcal{X}\\y \sim \mathcal{Y}}$}} \right)
    \end{equation}
\end{mydefinition}

% \paragraph{How to evaluate data distillation techniques?} 
From \cref{def:data_distillation_technical}, we highlight three cornerstones of evaluating data distillation methods: (1) Performance: downstream evaluation of models trained on the synthesized data summary \vs the full dataset (\eg, accuracy, FID, nDCG, \etc); (2) Efficiency: how quickly can models reach full-data performance (or even exceed it), \ie, the scaling of $n$ \vs downstream task-performance; and (3) Transferability: how well can data summaries generalize to a diverse pool of learning algorithms, in terms of downstream evaluation.

\paragraph{No free lunch.} The universal ``No Free Lunch'' theorem \citep{no_free_lunch} applies to data distillation as well. 
For example, looking at the transferability of a data summary, it
% which in our scenario translates to the fact that the transferability of a data summary 
is strongly dependent on the set of encoded inductive biases, \ie, through the choice of the learning algorithm $\Phi$ used while distilling, as well as the objective function $l(\cdot, \cdot)$. Such biases are unavoidable for any data distillation technique, 
in a sense that learning algorithms closely following the set of encoded inductive biases, will be able to generalize better on the data summary than others.

Keeping these preliminaries in mind, we now present a formal framework for data distillation, encapsulating existing data distillation approaches. Notably, the majority of existing techniques intrinsically solve a bilevel optimization problem, which are tractable surrogates of \cref{eqn:dd_optimization}. The inner-loop typically optimizes a representative learning algorithm on the data summary, and using the optimized learning algorithm, the outer-loop optimizes a tractable proxy of \cref{eqn:dd_optimization}. 

Some common assumptions that existing data distillation techniques follow are: (1) static-length data summary, \ie, $n$ is fixed and is treated as a tunable hyper-parameter; and (2) we have on-demand access to the target dataset $\dataset$ which is also assumed to be \texttt{iid}. Notably, the outer-loop optimization of \distill happens simply through GD on the analogously defined $\mathbf{X}_{\mathsf{syn}} \in \mathbb{R}^{n \times \dim(\mathcal{X})}$, which is instantiated as free parameters. Note that the labels, $\mathbf{Y}_{\mathsf{syn}} \in \mathbb{R}^{n \times \dim(\mathcal{Y})}$, can be similarly optimized through GD as well \citep{label_solve}. For the sake of notational clarity, we will interchangeably use optimization of \distill \emph{or} $(\mathbf{X}_{\mathsf{syn}}, \mathbf{Y}_{\mathsf{syn}})$ henceforth.

\subsection{Data Distillation by Meta-model Matching} \label{sec:meta_model}
Meta-model matching-based data distillation approaches fundamentally optimize for the transferability of models trained on the data summary when generalized to the original dataset:
\begin{equation} \label{eqn:meta_model_matching}
\begin{gathered}
    \underset{\distill}{\operatorname{arg} \operatorname{min}} \hspace{0.4cm} \mathcal{L}_{\dataset}\left(\theta^{\distill}\right) \hspace{0.3cm}
    \text{s.t.} \hspace{0.4cm} \theta^{\distill} ~ \triangleq ~ \underset{\theta}{\operatorname{arg} \operatorname{min}} \hspace{0.15cm} \mathcal{L}_{\distill}(\theta),
\end{gathered}
\end{equation}
where intuitively, the inner-loop trains a representative learning algorithm on the data summary \emph{until convergence}, and the outer-loop subsequently optimizes the data summary for the transferability of the optimized learning algorithm to the original dataset. 
Besides common assumptions mentioned earlier, the key simplifying assumption for this family of methods is that a perfect classifier exists and can be estimated on $\dataset$, \ie, $\exists ~\theta^{\dataset}$ s.t. $l(\Phi_{\theta^{\dataset}}(x), y) = 0, ~ \forall x \sim \mathcal{X}, y \sim \mathcal{Y}$. Plugging the second assumption along with the \texttt{iid} assumption of $\dataset$ in \cref{eqn:dd_optimization} directly translates to \cref{eqn:meta_model_matching}. Despite the assumption, \cref{eqn:meta_model_matching} is highly expensive both in terms of computation time and memory, due to which, methods from this family typically resort to making further assumptions.

\citet{dd_orig} (DD) originally proposed the task of data distillation, and used the meta-model matching framework for optimization. DD makes
% The first works on data distillation used the meta-model matching framework. \citet{dd_orig} (DD) make 
the optimization in \cref{eqn:meta_model_matching} tractable by performing (1) local optimization \emph{\`a la} stochastic gradient descent (SGD) in the inner-loop, and (2) outer-loop optimization using Truncated Back-Propagation Through Time (TBPTT), \ie, unrolling only a limited number of inner-loop optimization steps. Formally, the modified optimization objective for DD is as follows:
\begin{equation} \label{eqn:dd_orig_optimization}
\begin{gathered}
    \underset{\distill}{\operatorname{arg} \operatorname{min}} \hspace{0.4cm} \expectation{\theta_0 \sim \mathbf{P}_\theta}{\mathcal{L}_{\dataset}\left(\theta_T\right)} \hspace{0.3cm}
    \text{s.t.} \hspace{0.4cm} \theta_{t+1} \leftarrow \theta_t - \eta \cdot \nabla_\theta \mathcal{L}_{\distill}(\theta_t),
\end{gathered}
\end{equation}
where $\mathbf{P}_\theta$ is a parameter initialization distribution of choice, $T$ accounts for the truncation in TBPTT, and $\eta$ is a tunable learning rate. We also elucidate DD's control-flow in \cref{alg:dd_overview} for reference. 

Notably, TBPTT has been associated with drawbacks such as (1) computationally expensive inner-loop unrolling; (2) bias involved with truncated unrolling \citep{biased_bptt}; and (3) poorly conditioned loss landscapes, particularly with long unrolls \citep{bptt_loss_landscape}. Consequently, the TBPTT framework was empirically shown to be ineffective for data distillation \citep{zhao_dc}. However, recent work \citep{remember_past} claims that using momentum-based optimizers and longer inner-loop unrolling can greatly improve performance. We delay a deeper discussion of this work to \cref{sec:dd_factorization} for clarity.
% Missing work here

Analogously, a separate line of work focuses on using Neural Tangent Kernel (NTK) \citep{ntk} based algorithms to solve the inner-loop in closed form. As a brief side note, the infinite-width correspondence states that performing Kernelized Ridge Regression (KRR) using the NTK of a given neural network, is equivalent to training the same $\infty$-width neural network with L2 reconstruction loss for $\infty$ SGD-steps. These ``$\infty$-width'' neural networks have been shown to perform reasonably compared to their finite-width counterparts, while also being solved in closed-form (see \citet{finite_vs_infinite_2} for a detailed analysis on finite \vs infinite neural networks for image classification). KIP uses the NTK of a fully-connected neural network \citep{kip}, or a convolutional network \citep{kip_conv} in the inner-loop of \cref{eqn:meta_model_matching} for efficient data distillation. More formally, given the NTK $\mathcal{K} : \mathcal{X} \times \mathcal{X} \mapsto \mathbb{R}$ of a neural network architecture, KIP optimizes the following objective:
\begin{equation} \label{eqn:kip_optimization}
\begin{gathered}
    \underset{\mathbf{X}_{\mathsf{syn}}, \mathbf{Y}_{\mathsf{syn}}}{\operatorname{arg} \operatorname{min}} \hspace{0.4cm} \left\lVert \mathbf{Y} - \mathbf{K}_{\mathbf{X} \mathbf{X}_{\mathsf{syn}}} \cdot (\mathbf{K}_{\mathbf{X}_{\mathsf{syn}} \mathbf{X}_{\mathsf{syn}}} + \lambda I)^{-1} \cdot \mathbf{Y}_{\mathsf{syn}} \right\rVert^2,
\end{gathered}
\end{equation}
where $\mathbf{K}_{AB} \in \mathbb{R}^{|A|\times|B|}$ represents the gramian matrix of two sets $A$ and $B$, and whose $(i, j)^{\text{th}}$ element is defined by $\mathcal{K}(A_i, B_j)$. Although KIP doesn't impose any additional simplifications to the meta-model matching framework, 
it has an $\mathcal{O}(|\dataset| \cdot n \cdot \dim(\mathcal{X}))$ time and memory complexity,
%computing large NTK matrices like $\mathbf{K}_{\mathbf{X}_{\mathsf{syn}} \mathbf{X}_{\mathsf{syn}}}$ for complicated architectures and their inverse are highly compute intensive procedures,
limiting its scalability. Subsequently, RFAD \citep{rfad} proposes using (1) the light-weight Empirical Neural Network Gaussian Process (NNGP) kernel \citep{nngp} instead of the NTK; and (2) a classification loss (\eg, NLL) instead of the L2-reconstruction loss for the outer-loop to get $\mathcal{O}(n)$ time complexity while also having better performance. On a similar note, FRePO \citep{frepo} decouples the feature extractor and a linear classifier in $\Phi$, and alternatively optimizes (1) the data summary along with the classifier, and (2) the feature extractor. To be precise, let $f_{\theta} : \mathcal{X} \mapsto \mathcal{X}'$ be the feature extractor, $g_{\psi} : \mathcal{X}' \mapsto \mathcal{Y}$ be the linear classifier, s.t. $\Phi(x) \equiv g_{\psi}(f_{\theta}(x)) ~\forall x \in \mathcal{X}$; the optimization objective for FRePO can be written as:
\begin{equation} \label{eqn:frepo_optimization}
\begin{gathered}
    \underset{\mathbf{X}_{\mathsf{syn}}, \mathbf{Y}_{\mathsf{syn}}}{\operatorname{arg} \operatorname{min}} \hspace{0.4cm} 
    \expectation{\theta_0 \sim \mathbf{P}_\theta}{\sum_{t=0}^T ~ \left\lVert \mathbf{Y} - \mathbf{K}^{\theta_t}_{\mathbf{X} \mathbf{X}_{\mathsf{syn}}} \cdot (\mathbf{K}^{\theta_t}_{\mathbf{X}_{\mathsf{syn}} \mathbf{X}_{\mathsf{syn}}} + \lambda I)^{-1} \cdot \mathbf{Y}_{\mathsf{syn}} \right\rVert^2} \\
    \text{s.t.} \hspace{0.4cm} \theta_{t+1} \leftarrow \theta_t - \eta \cdot \expectation{(x, y) \sim \distill}{\nabla_\theta l(g_{\psi}(f_{\theta}(x)), y)} ~;~ \mathbf{K}^{\theta}_{\mathbf{X}_{\mathsf{syn}} \mathbf{X}_{\mathsf{syn}}} \triangleq f_{\theta_t}(\mathbf{X}_{\mathsf{syn}}) f_{\theta_t}(\mathbf{X}_{\mathsf{syn}})^T,
\end{gathered}
\end{equation}
where $T$ represents the number of inner-loop update steps for the feature extractor $f_\theta$. Notably, (1) a wide architecture for $f_\theta$ is crucial for distillation quality in FRePO; and (2) despite the bilevel optimization, FRePO is shown to be more scalable compared to KIP (\cref{eqn:kip_optimization}), while also being more generalizable.

% \newcommand\mycommfont[1]{#1}
% \SetCommentSty{mycommfont}
\setlength{\algomargin}{1.5em}
\begin{algorithm}[t]
    \caption{Control-flow of data distillation using na\"ive meta-matching (\cref{eqn:dd_orig_optimization})}
    \label{alg:dd_overview}
    \KwIn{Target dataset \dataset, outer-loop iterations $K$, parameter initialization distribution $\mathbf{P}_\theta$, inner-loop iterations $T$, inner-loop learning rate $\eta$, outer-loop learning rate $\eta_{\mathsf{syn}}$} % $\mathbf{w}_T$, $\mathbf{m}_T$, $\mathbf{v}_T$, $\gamma$, $\alpha$, $\epsilon$, $L(w, x)$, meta-objective $f(w)$
    \textbf{Initialize:} $(\mathbf{X}_{\mathsf{syn}}^{0}, \mathbf{Y}_{\mathsf{syn}}^{0}) \sim \dataset$ % $d\mathbf{m} \leftarrow 0$, $d\mathbf{x} \leftarrow 0$, $d\mathbf{w} \leftarrow \nabla_\mathbf{w} f(\mathbf{w}_T)$
    
    \For(\tcp*[f]{\textbf{Outer-loop:}\ optimize \distill}){$k = 1, \ldots, K$} %\Comment{Outer-loop: optimize \distill}
    {
        Initialize $\theta_0 \sim \mathbf{P}_\theta$
        
        \For(\tcp*[f]{\textbf{Inner-loop:}\ optimize $\Phi$ on $\distill^{k-1}$}){$t = 1, \ldots, T$} %\Comment{Inner-loop: optimize $\Phi$ on $\distill^{k-1}$}
        {
            $\theta_{t} \leftarrow \theta_{t-1} - \eta \cdot \nabla_\theta \mathcal{L}_{\distill^{k-1}}(\theta_{t-1})$
        }

        $\mathbf{X}_{\mathsf{syn}}^{k} \leftarrow \mathbf{X}_{\mathsf{syn}}^{k-1} - \eta_{\mathsf{syn}} \cdot \nabla_{\mathbf{X}_{\mathsf{syn}}} \mathcal{L}_{\dataset}(\theta_{T})$ \tcp*[f]{Update $\mathbf{X}_{\mathsf{syn}}$ by computing unrolled meta-gradient}
        
        $\mathbf{Y}_{\mathsf{syn}}^{k} \leftarrow \mathbf{Y}_{\mathsf{syn}}^{k-1} - \eta_{\mathsf{syn}} \cdot \nabla_{\mathbf{Y}_{\mathsf{syn}}} \mathcal{L}_{\dataset}(\theta_{T})$ \tcp*[f]{Update $\mathbf{Y}_{\mathsf{syn}}$ by computing unrolled meta-gradient}
    }
    \KwOut{$\distill^{K} \equiv (\mathbf{X}_{\mathsf{syn}}^{K}, \mathbf{Y}_{\mathsf{syn}}^{K})$}
\end{algorithm}

\subsection{Data Distillation by Gradient Matching} \label{sec:grad_matching}
\begin{figure*}[t!] \centering
    \centering
    \includegraphics[width=0.9\linewidth]{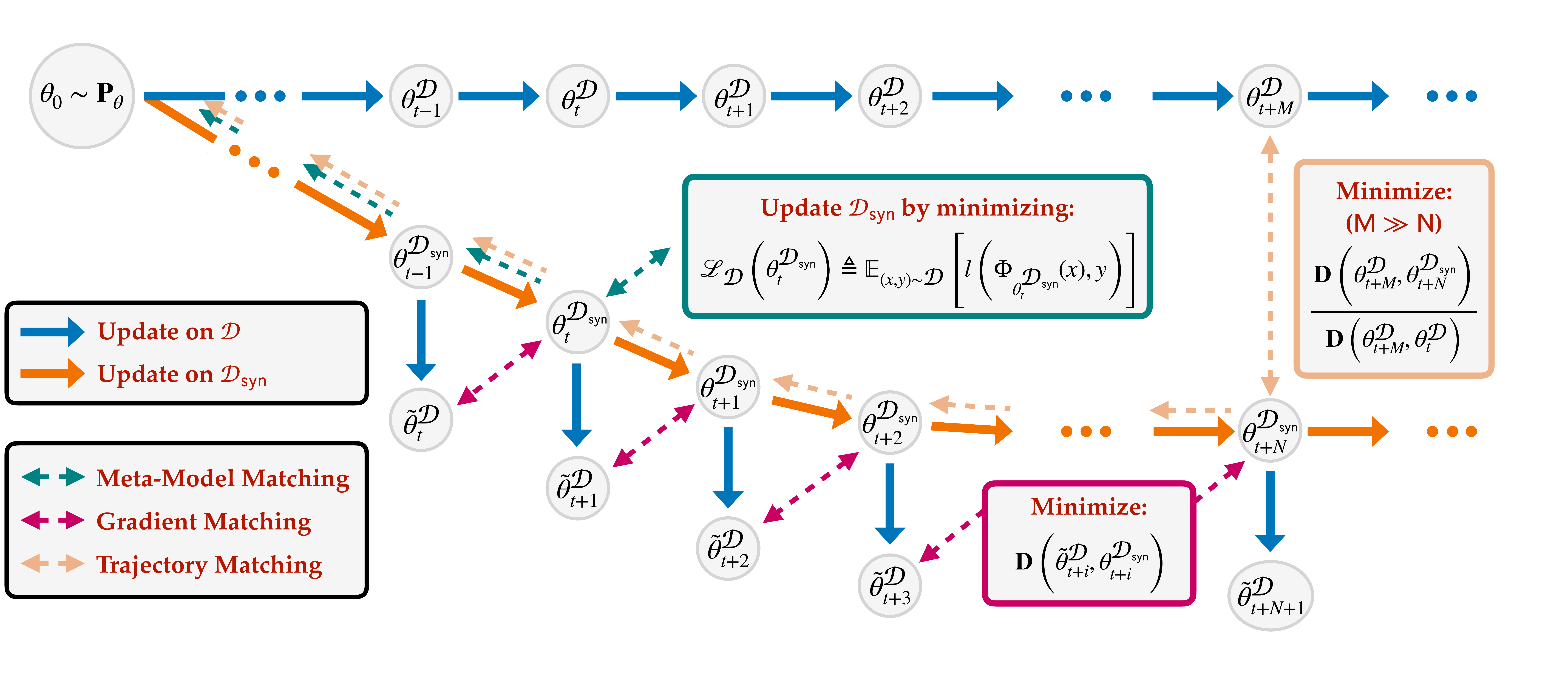}
    \renewcommand\figurename{\href{https://www.noveens.com/images/dd_survey/optimization.pdf}{[HQ Image Link]} Figure}
    \vspace{-10pt}
    \caption{The underlying optimization in various data distillation frameworks.}
    \label{fig:pictorial_gradients}
\end{figure*}

Gradient matching based data distillation, at a high level, performs one-step distance matching on (1) the network trained on the target dataset ($\dataset$) \vs (2) the same network trained on the data summary ($\distill$). In contrast to the meta-model matching framework, such an approach circumvents the unrolling of the inner-loop, thereby making the overall optimization much more efficient. 
% is a simplification of the aforementioned meta-model matching family to be more efficient. Approaches in this family aim to match a learning algorithm's gradient on the data summary ($\distill$) \vs on the target dataset ($\dataset$). 
First proposed by \citet{zhao_dc} (DC), data summaries optimized by gradient-matching significantly outperformed data pruning methodologies, as well as TBPTT-based data distillation proposed by \citet{dd_orig}. Formally, given a learning algorithm $\Phi$, DC solves the following optimization objective:
\begin{equation} \label{eqn:gradient_matching}
\begin{gathered}
    \underset{\distill}{\operatorname{arg} \operatorname{min}} \hspace{0.4cm} \expectation{\substack{\theta_0 \sim \mathbf{P}_\theta\\c ~\sim~ \mathcal{C}}}{\sum_{t=0}^T ~ \mathbf{D}\left( \nabla_\theta \mathcal{L}_{\dataset^c}(\theta_t), \nabla_\theta \mathcal{L}_{\distill^c}(\theta_t) \right)} \hspace{0.3cm}
    \text{s.t.} \hspace{0.4cm} \theta_{t+1} \leftarrow \theta_t - \eta \cdot \nabla_\theta \mathcal{L}_{\distill}(\theta_t),
\end{gathered}
\end{equation}
where $T$ accounts for model similarity $T$-steps in the future, and $\mathbf{D} : \mathbb{R}^{|\theta|} \times \mathbb{R}^{|\theta|} \mapsto \mathbb{R}$ is a distance metric of choice (typically cosine distance). 
In addition to assumptions imposed by the meta-model matching framework (\cref{sec:meta_model}), gradient-matching assumes (1) inner-loop optimization of only $T$ steps; (2) local smoothness: two sets of model parameters close to each other (given a distance metric) imply model similarity; and (3) first-order approximation of $\theta_t^{\dataset}$: instead of exactly computing the training trajectory of optimizing $\theta_0$ on $\dataset$ (say $\theta_t^{\dataset}$); perform first-order approximation on the optimization trajectory of $\theta_0$ on the much smaller $\distill$ (say $\theta_t^{\distill}$), \ie, approximate $\theta_t^{\dataset}$ as a single gradient-descent update on $\theta_{t-1}^{\distill}$ using $\dataset$ rather than $\theta_{t-1}^{\dataset}$ (\cref{fig:pictorial_gradients}).

% Provide a list of methods with their even subtler variations
Subsequently, numerous other approaches have been built atop this framework with subtle variations. DSA \citep{zhao_dsa} improves over DC by performing the same image-augmentations (\eg, crop, rotate, jitter, \etc) on both \dataset and \distill while optimizing \cref{eqn:gradient_matching}. Since these augmentations are universal and are applicable across data distillation frameworks, augmentations performed by DSA have become a common part of all methods proposed henceforth, but we omit them for notational clarity. DCC \citep{dcc} further modifies the gradient-matching objective to incorporate class contrastive signals inside each gradient-matching step and is shown to improve stability as well as performance. With $\theta_t$ evolving similarly as in \cref{eqn:gradient_matching}, the modified optimization objective for DCC can be written as:
\begin{equation} \label{eqn:dcc_gradient_matching}
\begin{gathered}
    \underset{\distill}{\operatorname{arg} \operatorname{min}} \hspace{0.4cm} \expectation{\theta_0 \sim \mathbf{P}_\theta}{\sum_{t=0}^T ~ \mathbf{D}\left( \expectation{c \in \mathcal{C}}{ \nabla_\theta \mathcal{L}_{\dataset^c}(\theta_t)}, \expectation{c \in \mathcal{C}}{\nabla_\theta \mathcal{L}_{\distill^c}(\theta_t)} \right)}
\end{gathered}
\end{equation}
Most recently, \citet{idc} (IDC) extend the gradient matching framework by: (1) multi-formation: to synthesize a higher amount of data within the same memory budget, store the data summary (\eg, images) in a lower resolution to remove spatial redundancies, and upsample (using \eg, bilinear, FSRCNN \citep{fsrcnn}) to the original scale while usage; and (2) matching gradients of the network's training trajectory over the full dataset $\dataset$ rather than the data summary $\distill$. To be specific, given a $k \times$ upscaling function $f : \mathbb{R}^{d \times d} \mapsto \mathbb{R}^{kd \times kd}$, the modified optimization objective for IDC can be formalized as:
\begin{equation} \label{eqn:idc_gradient_matching}
\begin{gathered}
    \underset{\distill}{\operatorname{arg} \operatorname{min}} \hspace{0.4cm} \expectation{\substack{\theta_0 \sim \mathbf{P}_\theta\\c ~\sim~ \mathcal{C}}}{\sum_{t=0}^T ~ \mathbf{D}\left( \nabla_\theta \mathcal{L}_{\dataset^c}(\theta_t), \nabla_\theta \mathcal{L}_{f(\distill^c)}(\theta_t) \right)} \hspace{0.3cm}
    \text{s.t.} \hspace{0.4cm} \theta_{t+1} \leftarrow \theta_t - \eta \cdot \nabla_\theta \mathcal{L}_{\dataset}(\theta_t)
\end{gathered}
\end{equation}
\citet{idc} further hypothesize that training models on $\distill$ instead of $\dataset$ in the inner-loop has two major drawbacks: (1) strong coupling of the inner- and outer-loop resulting in a chicken-egg problem \citep{em}; and (2) vanishing network gradients due to the small size of $\distill$, leading to an improper outer-loop optimization for gradient-matching based techniques.

\subsection{Data Distillation by Trajectory Matching} \label{sec:traj_matching}
\citet{mtt} proposed MTT which aims to match the training trajectories of models trained on \dataset \vs \distill. More specifically, let $\{ \theta_t^\dataset \}_{t=0}^T$ represent the training trajectory of training $\Phi_\theta$ on \dataset; trajectory matching algorithms aim to solve the following optimization:
\begin{equation} \label{eqn:traj_matching}
\begin{gathered}
    \underset{\distill, \eta}{\operatorname{arg} \operatorname{min}} \hspace{0.4cm} \expectation{\substack{\theta_0 \sim \mathbf{P}_\theta}}{\sum_{t=0}^{T-M} ~ \frac{\mathbf{D}\left( \theta_{t+M}^\dataset, \theta_{t+N}^{\distill} \right)}{\mathbf{D}\left( \theta_{t+M}^\dataset, \theta_{t}^{\dataset} \right)}} \\ % \hspace{0.3cm}
    \text{s.t.} \hspace{0.4cm} \theta_{t+i+1}^{\distill} \leftarrow \theta_{t+i}^{\distill} - \eta \cdot \nabla_\theta \mathcal{L}_{\distill}(\theta_{t+i}^{\distill}) \hspace{0.3cm};\hspace{0.3cm} \theta_{t+1}^{\distill} \leftarrow \theta_{t}^{\dataset} - \eta \cdot \nabla_\theta \mathcal{L}_{\distill}(\theta_{t}^{\dataset}) ,
\end{gathered}
\end{equation}
where $\mathbf{D} : \mathbb{R}^{|\theta|} \times \mathbb{R}^{|\theta|} \mapsto \mathbb{R}$ is a distance metric of choice (typically L2 distance). Such an optimization can intuitively be seen as optimizing for similar quality models trained with $N$ SGD steps on \distill, compared to $M \gg N$ steps on \dataset, thereby invoking long-horizon trajectory matching. Notably, calculating the gradient of \cref{eqn:traj_matching} \wrt \distill encompasses gradient unrolling through $N$-timesteps, thereby limiting the scalability of MTT. On the other hand, since the trajectory of training $\Phi_\theta$ on \dataset, \ie, $\{ \theta_t^\dataset \}_{t=0}^T$ is independent of the optimization of \distill, it can be pre-computed for various $\theta_0 \sim \mathbf{P}_\theta$ initializations and directly substituted. Similar to gradient matching methods (\cref{sec:grad_matching}), the trajectory matching framework also optimizes the first-order distance between parameters, thereby inheriting the local smoothness assumption. As a scalable alternative, \citet{tesla} proposed TESLA, which re-parameterizes the parameter-matching loss of MTT in \cref{eqn:traj_matching} (specifically when $\mathbf{D}$ is set as the L2 distance), using linear algebraic manipulations to make the bilevel optimization's memory complexity independent of $N$. Furthermore, TESLA uses learnable soft-labels ($\mathbf{Y}_{\mathsf{syn}}$) during the optimization for an increased compression efficiency.

\subsection{Data Distillation by Distribution Matching} \label{sec:distr_matching}
% \citep{IDC} for theory on why gradient matching is better than distribution matching
Even though the aforementioned gradient-matching or trajectory-matching based data distillation techniques have been empirically shown to synthesize high-quality data summaries, the underlying bilevel optimization, however, is oftentimes a computationally expensive procedure. To this end, distribution-matching techniques solve a correlated proxy task via a single-level optimization, leading to a vastly improved scalability. More specifically, instead of matching the quality of models on $\dataset$ \vs $\distill$, distribution-matching techniques directly match the distribution of $\dataset$ \vs $\distill$. The key assumption for this family of methods is that two datasets that are similar according to a particular distribution divergence metric, lead to similarly trained models.

First proposed by \citet{dm}, DM uses (1) numerous parametric encoders to cast high-dimensional data into respective low-dimensional latent spaces; and (2) an approximation of the Maximum Mean Discrepancy to compute the distribution mismatch between $\dataset$ and $\distill$ in each of the latent spaces. More precisely, given a set of $k$ encoders $\mathcal{E} \triangleq \{ \psi_i : \mathcal{X} \mapsto \mathcal{X}_i \}_{i=1}^{k}$, the optimization objective can be written as:
\begin{equation} \label{eqn:distribution_matching}
\begin{gathered}
    \underset{\distill}{\operatorname{arg} \operatorname{min}} \hspace{0.4cm} \expectation{\substack{\psi \sim \mathcal{E}\\c ~\sim~ \mathcal{C}}}{\left\lVert \expectation{x \sim \dataset^c}{\psi(x)} - \expectation{x \sim \distill^c}{\psi(x)} \right\rVert^2}
\end{gathered}
\end{equation}
DM uses a set of randomly initialized neural networks (with the same architecture) to instantiate $\mathcal{E}$. They observe similar performance when instantiated with more meaningful, task-optimized neural networks, despite it being much less efficient. CAFE \citep{cafe} further refines the distribution-matching idea by: (1) solving a bilevel optimization problem for jointly optimizing a \emph{single} encoder ($\Phi$) and the data summary, rather than using a pre-determined \emph{set} of encoders ($\mathcal{E}$); and (2) assuming a neural network encoder ($\Phi$), match the latent representations obtained at all intermediate layers of the encoder instead of only the last layer. Formally, given a $(L+1)$-layer neural network $\Phi_\theta : \mathcal{X} \mapsto \mathcal{Y}$ where $\Phi^l_{\theta}$ represents $\Phi$'s output at the $l^{\text{th}}$ layer, the optimization problem for CAFE can be specified as:
\begin{equation} \label{eqn:cafe_distribution_matching}
\begin{gathered}
    \underset{\distill}{\operatorname{arg} \operatorname{min}} \hspace{0.4cm} \expectation{c ~\sim~ \mathcal{C}}{\sum_{l=1}^{L} \left\lVert \expectation{x \sim \dataset^c}{\Phi^l_{\theta_t}(x)} - \expectation{x \sim \distill^c}{\Phi^l_{\theta_t}(x)} \right\rVert^2  - \beta \cdot \expectation{(x,y)\sim \dataset^c}{\log \hat{p}(y|x, \theta_t)}} \\
    \text{s.t.} \hspace{0.4cm} \theta_{t+1} \leftarrow \theta_t - \eta \cdot \nabla_\theta \mathcal{L}_{\distill}(\theta_t) 
    \hspace{0.2cm} ; \hspace{0.2cm}
    \hat{p}(y|x, \theta) \triangleq \underset{y}{\operatorname{softmax}}\left(\left\langle \Phi^L_\theta(x), \expectation{x' \sim \distill^y}{\Phi^L_\theta(x')} \right\rangle\right),
\end{gathered}
\end{equation}
where $\hat{p}(\cdot|\cdot, \theta)$ intuitively represents the nearest centroid classifier on $\distill$ using the latent representations obtained by last layer of $\Phi_\theta$. Analogously, IT-GAN \citep{gan_distillation} also uses the distribution-matching framework in \cref{eqn:distribution_matching} to generate data that is informative for model training, in contrast to the traditional GAN \citep{gan} which focuses on generating realistic data.

\newcommand{\format}[2]{\begin{tabular}{@{}c@{}}$#1$\\[-7pt]$\scriptscriptstyle \pm #2$\end{tabular}}
\newcommand{\boformat}[2]{\color{black} \begin{tabular}{@{}c@{}}$\mathbf{#1}$\\[-7pt]$\scriptscriptstyle \mathbf{\pm #2}$\end{tabular}}
\newcommand{\orformat}[2]{\color{orange} \begin{tabular}{@{}c@{}}$\mathbf{#1}$\\[-7pt]$\scriptscriptstyle \mathbf{\pm #2}$\end{tabular}}
\newcommand{\blformat}[2]{\color{blue} \begin{tabular}{@{}c@{}}$\mathbf{#1}$\\[-7pt]$\scriptscriptstyle \mathbf{\pm #2}$\end{tabular}}

\def\arraystretch{1.32}
\setlength{\tabcolsep}{0.5em}
\begin{table*}[!htbp]
    \begin{footnotesize}
    \begin{center}
        \caption{Comparison of data distillation methods. Each method (1) synthesizes the data summary on the train-set; (2) unless mentioned, trains a 128-width ConvNet \citep{conv_net} on the data summary; and (3) evaluates it on the test-set. Confidence intervals are obtained by training at least 5 networks on the data summary. LinBa (No Fact.) represents LinBa with the no factorization. Methods evaluated using KRR are marked as ($\infty$-Conv) or ($\infty$-FC). The equivalent storage-in-bytes is used for factorization-based techniques instead of IPC. The best method in their category is \textbf{emboldened}, the best-overall non-factorized method evaluated on ConvNet is \textbf{\textcolor{orange}{colored orange}}, and the best-overall factorized method is \textbf{\textcolor{blue}{colored blue}}.}
        \label{tab:main_results}
        \begin{tabular}{c c | c c c | c c c | c c c | c c c }
            \toprule
            \multicolumn{2}{c|}{\textbf{Dataset}} & \multicolumn{3}{c|}{\textbf{MNIST}} & \multicolumn{3}{c|}{\textbf{CIFAR-10}} & \multicolumn{3}{c|}{\textbf{CIFAR-100}} & \multicolumn{3}{c}{\textbf{Tiny ImageNet}} \\
            \multicolumn{2}{c|}{\textbf{Imgs/Class (IPC)}} & 1 & 10 & 50 & 1 & 10 & 50 & 1 & 10 & 50 & 1 & 10 & 50 \\
            
            \midrule
            
            \multirow{4}{*}{\STAB{\rotatebox[origin=c]{90}{\begin{tabular}{@{}c@{}}Baselines\\\end{tabular}}}} & Random & \format{64.9}{3.5} & \boformat{95.1}{0.9} & \boformat{97.9}{0.2} & \format{14.4}{2.0} & \format{26.0}{1.2} & \boformat{43.4}{1.0} & \format{4.2}{0.3} & \format{14.6}{0.5} & \format{30.0}{0.4} & \boformat{1.5}{0.1} & \boformat{6.0}{0.8} & \boformat{16.8}{1.8} \\ 
            & Herding\color{red}{$^1$} & \boformat{89.2}{1.6} & \format{93.7}{0.3} & \format{94.9}{0.2} & \boformat{21.5}{1.2} & \boformat{31.6}{0.7} & \format{40.4}{0.6} & \format{8.4}{0.3} & \format{17.3}{0.5} & \format{33.7}{0.5} & - & - & - \\ 
            & Forgetting\color{red}{$^2$} & \format{35.5}{5.6} & \format{68.1}{3.3} & \format{88.2}{1.2} & \format{13.5}{1.2} & \format{23.3}{1.0} & \format{23.3}{1.1} & \format{4.5}{0.2} & \format{15.1}{0.3} & \format{30.5}{0.3} & - & - & - \\ 
            
            \midrule

            \multirow{12}{*}{\STAB{\rotatebox[origin=c]{90}{\begin{tabular}{@{}c@{}}Meta-model Matching\\\end{tabular}}}} & DD\color{red}{$^3$} & - & \format{79.5}{8.1} & - & - & \format{36.8}{1.2} & - & - & - & - & - & - & - \\
            & LinBa (No Fact.)\color{red}{$^{16}$} & \orformat{95.2}{0.3} & \orformat{98.8}{0.1} & \orformat{99.2}{0.1} & \format{49.1}{0.6} & \format{62.4}{0.4} & \format{70.5}{0.4} & \format{21.3}{0.6} & \format{34.7}{0.5} & - & - & - & - \\
            & KIP (ConvNet)\color{red}{$^4$} & \format{90.1}{0.1} & \format{97.5}{0.0} & \format{98.3}{0.1} & \format{49.9}{0.2} & \format{62.7}{0.3} & \format{68.6}{0.2} & \format{15.7}{0.2} & \format{28.3}{0.1} & - & - & - & - \\
            & RFAD (ConvNet)\color{red}{$^5$} & \format{94.4}{1.5} & \format{98.5}{0.1} & \format{98.8}{0.1} & \orformat{53.6}{1.2} & \boformat{66.3}{0.5} & \format{71.1}{0.4} & \format{26.3}{1.1} & \format{33.0}{0.3} & - & - & - & - \\
            & FRePO (ConvNet)\color{red}{$^6$} & \format{93.0}{0.4} & \format{98.6}{0.1} & \orformat{99.2}{0.1} & \format{46.8}{0.7} & \format{65.5}{0.6} & \boformat{71.7}{0.2} & \orformat{28.7}{0.1} & \orformat{42.5}{0.2} & \boformat{44.3}{0.2} & \orformat{15.4}{0.3} & \orformat{25.4}{0.2} & - \\

            \cmidrule{2-14}
            
            & KIP ($\infty$-FC)\color{red}{$^7$} & \format{85.5}{0.1} & \format{97.2}{0.2} & \format{98.4}{0.1} & \format{40.5}{0.4} & \format{53.1}{0.5} & \format{58.6}{0.4} & - & - & - & - & - & - \\
            & KIP ($\infty$-Conv)\color{red}{$^4$} & \boformat{97.3}{0.1} & \boformat{99.1}{0.1} & \boformat{99.5}{0.1} & \boformat{64.7}{0.2} & \boformat{75.6}{0.2} & \boformat{80.6}{0.1} & \format{34.9}{0.1} & \boformat{49.5}{0.3} & - & - & - & - \\
            & RFAD ($\infty$-Conv)\color{red}{$^5$} & \format{97.2}{0.2} & \format{99.1}{0.0} & \format{99.1}{0.0} & \format{61.4}{0.8} & \format{73.7}{0.2} & \format{76.6}{0.3} & \boformat{44.1}{0.1} & \format{46.8}{0.2} & - & - & - & - \\
            & FRePO ($\infty$-Conv)\color{red}{$^6$} & \format{92.6}{0.4} & \format{98.6}{0.1} & \format{99.2}{0.1} & \format{47.9}{0.6} & \format{68.0}{0.2} & \format{74.4}{0.1} & \format{32.3}{0.1} & \format{44.9}{0.2} & \format{43.0}{0.3} & \boformat{19.1}{0.3} & \boformat{26.5}{0.1} & - \\

            \midrule

            \multirow{4}{*}{\STAB{\rotatebox[origin=c]{90}{\begin{tabular}{@{}c@{}}Gradient\\Matching\\\end{tabular}}}} & DC\color{red}{$^8$} & \boformat{91.7}{0.5} & \format{97.4}{0.2} & \format{98.2}{0.2} & \format{28.3}{0.5} & \format{44.9}{0.5} & \format{53.9}{0.5} & \format{12.8}{0.3} & \format{25.2}{0.3} & \format{30.5}{0.3} & \format{4.6}{0.6} & \format{11.2}{1.6} & \format{10.9}{0.7} \\
            & DSA\color{red}{$^9$} & \format{88.7}{0.6} & \boformat{97.8}{0.1} & \orformat{99.2}{0.1} & \format{28.8}{0.7} & \format{52.1}{0.5} & \format{60.6}{0.5} & \format{13.9}{0.3} & \format{32.3}{0.3} & \boformat{42.8}{0.4} & \boformat{6.6}{0.2}  & \boformat{14.4}{2.0} & \boformat{22.6}{2.6} \\
            & DCC\color{red}{$^{10}$} & - & - & - & \boformat{34.0}{0.7} & \boformat{54.5}{0.5} & \boformat{64.2}{0.4} & \boformat{14.6}{0.3} & \boformat{33.5}{0.3} & \format{39.3}{0.4} & - & - & - \\

            \midrule

            % \multirow{2}{*}{\STAB{\rotatebox[origin=c]{90}{\begin{tabular}{@{}c@{}}Distr.\\Matching\\\end{tabular}}}} & DM\color{red}{$^{11}$} & \format{64.9}{3.5} & \format{95.1}{0.9} & \format{97.9}{0.2} & \format{14.4}{2.0} & \format{26.0}{1.2} & \format{43.4}{1.0} & \format{4.2}{0.3} & \format{14.6}{0.5} & \format{30.0}{0.4} & \format{3.9}{0.2} & \format{12.9}{0.4} & \format{21.5}{1.2} \\
            \multirow{2}{*}{\begin{tabular}{@{}c@{}}\\[-10pt]Distr.\\Matching\end{tabular}} & DM\color{red}{$^{11}$} & \format{89.7}{0.6} & \boformat{97.5}{0.1} & \format{98.6}{0.1} & \format{26.0}{0.8} & \format{48.9}{0.6} & \boformat{63.0}{0.4} & \format{11.4}{0.3} & \format{29.7}{0.3} & \boformat{43.6}{0.4} & \boformat{3.9}{0.2} & \boformat{12.9}{0.4} & \boformat{24.1}{0.3} \\
            & CAFE\color{red}{$^{12}$} & \boformat{90.8}{0.5} & \boformat{97.5}{0.1} & \boformat{98.9}{0.2} & \boformat{31.6}{0.8} & \boformat{50.9}{0.5} & \format{62.3}{0.4} & \boformat{14.0}{0.3} & \boformat{31.5}{0.2} & \format{42.9}{0.2} & - & - & - \\

            \midrule

            \multirow{2}{*}{\begin{tabular}{@{}c@{}}\\[-10pt]Traj.\\Matching\end{tabular}}
            % \begin{tabular}{@{}c@{}}Traj.\\Matching\end{tabular} 
            & MTT\color{red}{$^{13}$} & - & - & - & \format{46.3}{0.8} & \format{65.3}{0.7} & \format{71.6}{0.2} & \format{24.3}{0.3} & \format{40.1}{0.4} & \format{47.7}{0.2} & \boformat{8.8}{0.3} & \boformat{23.2}{0.2} & \orformat{28.0}{0.3} \\
            & TESLA\color{red}{$^{14}$} & - & - & - & \boformat{48.5}{0.8} & \orformat{66.4}{0.8} & \orformat{72.6}{0.7} & \boformat{24.8}{0.4} & \boformat{41.7}{0.3} & \orformat{47.9}{0.3} & - & - & - \\
            % \multirow{1}{*}{\STAB{\rotatebox[origin=c]{90}{\begin{tabular}{@{}c@{}}Traj.\\Matching\\\end{tabular}}}} & MTT\color{red}{$^{13}$} & \format{64.9}{3.5} & \format{95.1}{0.9} & \format{97.9}{0.2} & \format{46.3}{0.8} & \format{65.3}{0.7} & \format{71.6}{0.2} & \format{24.3}{0.3} & \format{40.1}{0.4} & \format{47.7}{0.2} & \format{8.8}{0.3} & \format{23.2}{0.2} & \format{28.0}{0.3} \\

            \midrule

            \multirow{5}{*}{\STAB{\rotatebox[origin=c]{90}{\begin{tabular}{@{}c@{}}Factorization \ \\\end{tabular}}}} & IDC\color{red}{$^{15}$} & - & - & - & \format{50.0}{0.4} & \format{67.5}{0.5} & \format{74.5}{0.1} & - & \format{44.8}{0.2} & - & - & - & - \\
            & LinBa\color{red}{$^{16}$} & \blformat{98.7}{0.7} & \blformat{99.3}{0.5} & \blformat{99.4}{0.4} & \blformat{66.4}{0.4} & \format{71.2}{0.4} & \format{73.6}{0.5} & \format{34.0}{0.4} & \format{42.9}{0.7} & - & \format{16.0}{0.7}  & - & - \\
            & HaBa\color{red}{$^{17}$} & - & - & - & \format{48.3}{0.8} & \format{69.9}{0.4} & \format{74.0}{0.2} & \format{33.4}{0.4} & \format{40.2}{0.2} & \blformat{47.0}{0.2} & - & - & - \\
            & KFS\color{red}{$^{18}$} & - & - & - & \format{59.8}{0.5} & \blformat{72.0}{0.3} & \blformat{75.0}{0.2} & \blformat{40.0}{0.5} & \blformat{50.6}{0.2} & - & \blformat{22.7}{0.2} & \blformat{27.8}{0.2} & - \\

            \midrule

            \multicolumn{2}{c|}{Full Dataset} & \multicolumn{3}{c|}{\format{99.6}{0.1}} & \multicolumn{3}{c|}{\format{84.8}{0.1}} & \multicolumn{3}{c|}{\format{56.2}{0.3}} & \multicolumn{3}{c}{\format{37.6}{0.4}} \\
            
            \bottomrule

            \\[-5pt]

            \multicolumn{14}{l}{\textcolor{red}{$^1$} \citep{herding}, \textcolor{red}{$^2$} \citep{forgetting}, \textcolor{red}{$^3$} \citep{dd_orig}, \textcolor{red}{$^4$} \citep{kip_conv}, \textcolor{red}{$^5$} \citep{rfad}} \\
            \multicolumn{14}{l}{\textcolor{red}{$^6$} \citep{frepo}, \textcolor{red}{$^7$} \citep{kip}, \textcolor{red}{$^8$} \citep{zhao_dc}, \textcolor{red}{$^9$} \citep{zhao_dsa}, \textcolor{red}{$^{10}$} \citep{dcc}} \\
            \multicolumn{14}{l}{\textcolor{red}{$^{11}$} \citep{dm}, \textcolor{red}{$^{12}$} \citep{cafe}, \textcolor{red}{$^{13}$} \citep{mtt}, \textcolor{red}{$^{14}$} \citep{tesla}} \\
            \multicolumn{14}{l}{\textcolor{red}{$^{15}$} \citep{idc}, \textcolor{red}{$^{16}$} \citep{remember_past}, \textcolor{red}{$^{17}$} \citep{haba}, \textcolor{red}{$^{18}$} \citep{kfs}} \\
        \end{tabular}
    \end{center}
    \end{footnotesize}
    % \bigskip
    \vspace{-6mm} %Put here to reduce too much white space after your table
\end{table*}

\subsection{Data Distillation by Factorization} \label{sec:dd_factorization}
All of the aforementioned data distillation frameworks intrinsically maintain the synthesized data summary as a large set of free parameters, which are in turn optimized. Arguably, such a setup prohibits knowledge sharing between synthesized data points (parameters), which might introduce data redundancy. On the other hand, factorization-based data distillation techniques parameterize the data summary using two separate components: (1) bases: a set of mutually independent base vectors; and (2) hallucinators: a mapping from the bases' vector space to the joint data- and label-space. In turn, both the bases and hallucinators are optimized for the task of data distillation. 

Formally, let $\mathcal{B} \triangleq \{b_i \in \mathbb{B}\}_{i=1}^{|\mathcal{B}|}$ be the set of bases, and $\mathcal{H} \triangleq \{h_i : \mathbb{B} \mapsto \mathcal{X} \times \mathcal{Y} \}_{i=1}^{|\mathcal{H}|}$ be the set of hallucinators, then the data summary is parameterized as $\distill \triangleq \{ {h(b)} \}_{b\sim \mathcal{B}, ~ h\sim \mathcal{H}}$. Even though such a two-pronged approach seems similar to generative modeling of data, note that unlike classic generative models, (1) the input space consists \emph{only of} a fixed and optimized set of latent codes and isn't meant to take any other inputs; and (2) given a specific $\mathcal{B}$ and $\mathcal{H}$, we can generate at most $|\mathcal{B}|\cdot|\mathcal{H}|$ sized data summaries. Notably, such a hallucinator-bases data parameterization can be optimized using any of the aforementioned data optimization frameworks (\cref{sec:meta_model,sec:grad_matching,sec:distr_matching,sec:traj_matching})

This framework was concurrently proposed by \citet{remember_past} (we take the liberty to term their unnamed model as ``\emph{Lin}-ear \emph{Ba}-ses'') and \citet{haba} (HaBa). LinBa modifies the general hallucinator-bases framework by assuming (1) the bases' vector space ($\mathbb{B}$) to be the same as the task input space ($\mathcal{X}$); and (2) the hallucinator to be linear and additionally conditioned on a given predictand. More specifically, the data parameterization can be formalized as follows:
\begin{equation} \label{eqn:linba_generation}
\begin{gathered}
    \distill \triangleq \left\{ ~ (y \ \mathbf{H}^T\mathbf{B}, ~y) ~ \right\}_{\raisebox{6pt}{$\substack{y \sim \mathcal{C}\\\mathbf{H} \sim \mathcal{H}}$}} \\
    \text{s.t.} \hspace{0.4cm}
    \mathbf{B} \in \mathbb{R}^{|\mathbf{B}| \times \dim(\mathcal{X})} \triangleq \left[ b_i \in \mathcal{X} \right]_{i=1}^{|\mathbf{B}|}
    \hspace{0.3cm} ; \hspace{0.3cm} 
    \mathcal{H} \triangleq \left\{ \mathbf{H}_i \in \mathbb{R}^{|\mathbf{B}| \times |\mathcal{C}|} \right\}_{i=1}^{|\mathcal{H}|},
\end{gathered}
\end{equation}
where for the sake of notational simplicity, we assume $y \in \mathbb{R}^{|\mathcal{C}|}$ represents the one-hot vector of the label for which we want to generate data, and the maximum amount of data that can be synthesized $n \leq |\mathcal{C}|\cdot|\mathcal{H}|$. Since the data generation (\cref{eqn:linba_generation}) is end-to-end differentiable, both $\mathbf{B}$ and $\mathcal{H}$ are jointly optimized using the TBPTT framework discussed in \cref{sec:meta_model}, albeit with some crucial modifications for vastly improved performance: (1) using momentum-based optimizers instead of vanilla SGD in the inner-loop; and (2) longer unrolling ($\geq 100$ steps) of the inner-loop during TBPTT. \citet{haba} (HaBa) relax the linear and predictand-conditional hallucinator assumption of LinBa, equating to the following data parameterization:
\begin{equation} \label{eqn:haba_generation}
\begin{gathered}
    \distill \triangleq \left\{ ~ (h(b), ~y) ~ \right\}_{\raisebox{6pt}{$\substack{b, y \sim \mathcal{B}\\h \sim \mathcal{H}}$}}
    \hspace{0.3cm} \text{s.t.} \hspace{0.4cm}
    \mathcal{B} \triangleq \left\{ ~ (b_i \in \mathcal{X}, y_i \in \mathcal{Y}) ~ \right\}_{i=1}^{|\mathcal{B}|}
    \hspace{0.3cm} ; \hspace{0.3cm} 
    \mathcal{H} \triangleq \left\{ h_{\theta_i} : \mathcal{X} \mapsto \mathcal{X} \right\}_{i=1}^{|\mathcal{H}|},
\end{gathered}
\end{equation}
where $\mathcal{B}$ and $\mathcal{H}$ are optimized using the trajectory matching framework (\cref{sec:traj_matching}) with an additional contrastive constraint to promote diversity in \distill (cf.~\citet{haba}, Equation (6)). Following this setup, HaBa can generate at most $|\mathcal{B}|\cdot|\mathcal{H}|$ sized data summaries. Furthermore, one striking difference between HaBa (\cref{eqn:haba_generation}) and LinBa (\cref{eqn:linba_generation}) is that to generate each data point, LinBa uses a linear combination of \emph{all} the bases, whereas HaBa generates a data point using a \emph{single} base vector. 

\citet{kfs} (KFS) further build atop this framework by maintaining a different bases' vector space $\mathbb{B}$ from the data domain $\mathcal{X}$, such that $\dim(\mathbb{B}) < \dim(\mathcal{X})$. This parameterization allows KFS to store an even larger number of images, with a comparable storage budget to other methods. Formally, the data parameterization for KFS can be specified as:
\begin{equation} \label{eqn:kfs_generation}
\begin{gathered}
    \distill \triangleq \bigcup_{c \in \mathcal{C}} \left\{ ~ (h(b), ~c) ~ \right\}_{\raisebox{6pt}{$\substack{b \sim \mathcal{B}_c\\h \sim \mathcal{H}}$}} \\
    \text{s.t.} \hspace{0.4cm}
    \mathcal{B} \triangleq \bigcup_{c \in \mathcal{C}} \mathcal{B}_c 
    \hspace{0.3cm} ; \hspace{0.3cm} 
    \mathcal{B}_c \triangleq \left\{ b^c_i \in \mathbb{B} \right\}_{i=1}^{B}
    \hspace{0.3cm} ; \hspace{0.3cm} 
    \mathcal{H} \triangleq \left\{ h_{\theta_i} : \mathbb{B} \mapsto \mathcal{X} \right\}_{i=1}^{|\mathcal{H}|},
\end{gathered}
\end{equation}
where KFS stores $B$ bases per class, equivalent to a total of $n = |\mathcal{C}| \cdot B \cdot |\mathcal{H}|$ sized data summaries. Following this data parameterization, $\mathcal{B}$ and $\mathcal{H}$ are optimized using the distribution matching framework for data distillation (\cref{eqn:distribution_matching}) to ensure fast, single-level optimization.

\paragraph{Data Distillation \vs Data Compression.} We highlight that it is non-trivial to ensure a fair comparison between data distillation techniques that (1) are ``non-factorized'', \ie, maintain each synthesized data point as a set of free-parameters (\cref{sec:meta_model,sec:grad_matching,sec:distr_matching,sec:traj_matching}); and (2) use factorized approaches discussed in this section to efficiently organize the data summary. If we use the size of the data summary ($n$) as the efficiency metric, factorized approaches are adversely affected as they need a much smaller storage budget to synthesize the same-sized data summaries. On the other hand, if we use ``end-to-end bytes of storage'' as the efficiency metric, non-factorized approaches are adversely affected as they perform no kind of data compression, but focus solely on better understanding the model-to-data relationship through the lens of optimization. For a better intuition, one can apply posthoc lossless compression (\eg, Huffman coding) on data synthesized by non-factorized data distillation approaches to fit more images in the same storage budget \citep{less_is_more}. Such techniques 
% More concerningly, in this setup, techniques 
unintentionally deviate from the original intent of data distillation, and progress more toward better data compression techniques. 
%To recommend a solution, 
As a potential solution,
we encourage the community to consider reporting results for both scenarios: a fixed data summary size $n$, as well as fixed bytes-of-storage. Nonetheless, for the ease of empirical comparison amongst the discussed data distillation techniques, we provide a collated set of results over four image-classification datasets in \cref{tab:main_results}.

\section{Data Modalities} \label{sec:data_modalities}
Having learned about different kinds of optimization frameworks for data distillation, we now discuss an orthogonal (and important) aspect of data distillation --- \emph{what kinds of data can data distillation techniques summarize?} From continuous-valued images to heterogeneous and semi-structured graphs, the underlying data for each unique application of machine learning has its own modality, structure, and set of assumptions. While the earliest data distillation techniques were designed to summarize images, recent steps have been taken to expand the horizon of data distillation into numerous other scenarios. In what follows, we categorize existing data distillation methods by their intended data modality, while also discussing their unique challenges.

\paragraph{Images.} A large-portion of existing data distillation techniques are designed for image classification data \citep{dd_orig, zhao_dc, zhao_dsa, kip, kip_conv, dm, idc, mtt, cafe, gan_distillation, frepo, rfad, dcc, haba, remember_past, kfs} simply because images have a real-valued, continuous data-domain ($\mathcal{X} \equiv \mathbb{R}^{d \times d}$). This allows SGD-based optimization directly on the data, which is treated as a set of free parameters. Intuitively, incrementally changing each pixel value can be treated as slight perturbations in the color space, and hence given a suitable data distillation loss, can be na\"ively optimized using SGD. 

\paragraph{Text.} Textual data is available in large amounts from sources like websites, news articles, academic manuscripts, \etc, and is also readily accessible with datasets like the common crawl\footnote{\href{https://commoncrawl.org/the-data/}{https://commoncrawl.org/the-data/}} which sizes up to almost 541TB. Furthermore, with the advent of large language models (LLM) \citep{bert, gpt, lamda}, training such models from scratch on large datasets has become an increasingly expensive procedure. Despite recent efforts in democratizing LLM training \citep{cramming, huggingface, bloom}, effectively distilling large-scale textual data as a solution is yet to be explored. The key bottlenecks for distilling textual data are: (1) the inherently discrete nature of data, where a token should belong in a limited vocabulary of words; (2) the presence of a rich underlying structure, \ie, sentences of words (text) obey fixed patterns according to a grammar; and (3) richness of context, \ie, a given piece of text could have wildly different semantic interpretations under different contexts.

\citet{text_distill} take a latent-embedding approach to textual data distillation. On a high level, to circumvent the discreteness of the optimization, the authors perform distillation in a continuous embedding space. More specifically, assuming access to a latent space specified by a \emph{fixed} text-encoder, the authors learn continuous \emph{representations} of each word in the distilled text and optimize it using the TBPTT data-distillation framework proposed by \citet{dd_orig} (\cref{eqn:dd_orig_optimization}). Finally, the distilled text representations are decoded by following a simple nearest-neighbor protocol.

\begin{figure*}[t!] \centering
    \centering
    \includegraphics[width=0.8\linewidth]{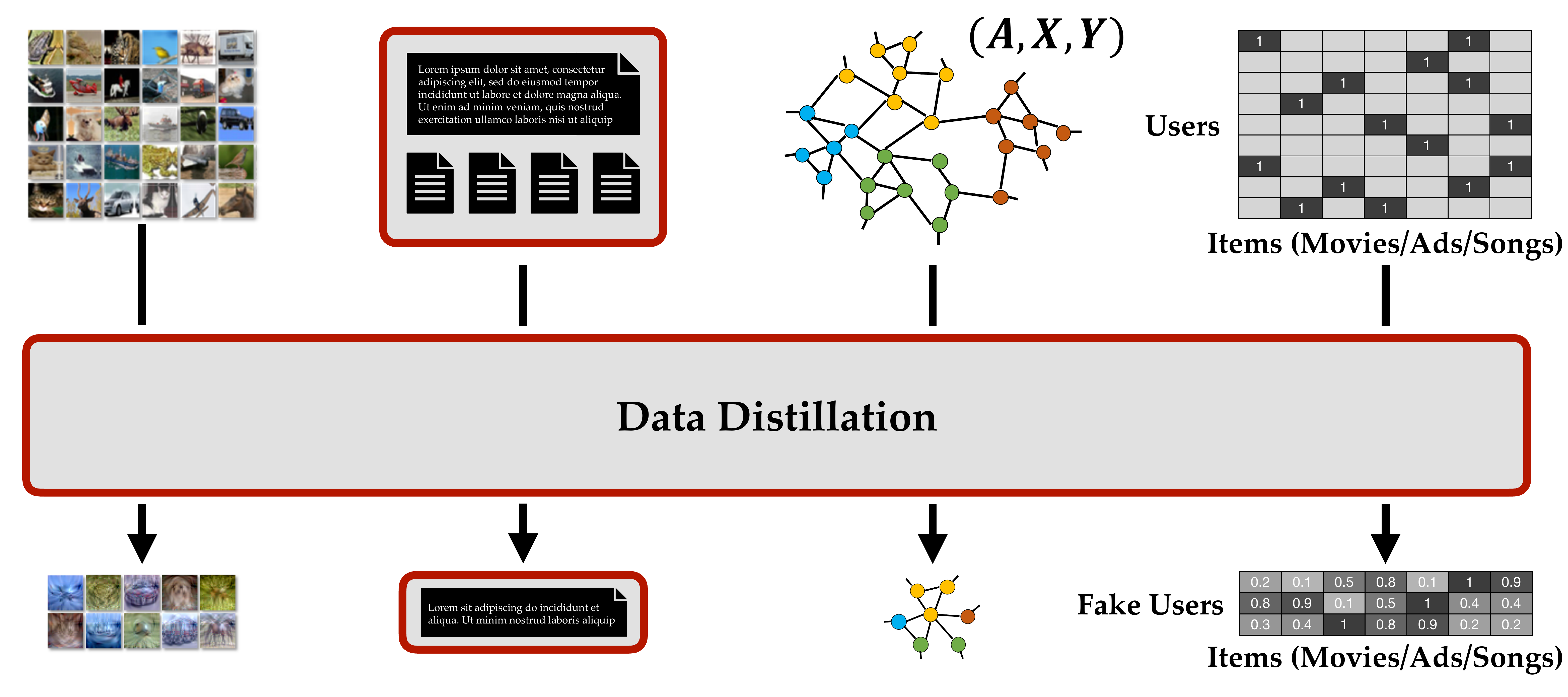}
    \renewcommand\figurename{\href{https://www.noveens.com/images/dd_survey/data_modalities.pdf}{[HQ Image Link]} Figure}
    \caption{Overview of distilling data for a few commonly observed data modalities.}
    \label{fig:data_modalities}
    \vspace{-10pt}
\end{figure*}

\paragraph{Graphs.} A wide variety of data and applications can inherently be modeled as graphs, \eg, user-item interactions \citep{gnn_recsys_survey, eclare, reviews_sigir}, social networks \citep{gnn_social}, autonomous driving \citep{gnn_self_driving, gapformer}, \etc Taking the example of social networks, underlying user-user graphs typically size up to the billion-scale \citep{graph_billion}, calling for principled scaling solutions. Graph distillation trivially solves a majority of the scale challenges, but synthesizing tiny, high-fidelity graphs has the following hurdles: (1) nodes in a graph can be highly abstract, \eg, users, products, \etc and could be discrete, heterogeneous, or even numerical IDs; (2) graphs follow a variety of intrinsic patterns (\eg, spatial \citep{gcn}) which need to be retained in the distilled graphs; and (3) quadratic size of the adjacency matrix could be computationally prohibitive for data optimization.

\citet{graph_distill_iclr_22} propose \textsc{GCond} which distills graphs in the inductive node-classification setting, specified by its node-feature matrix $\mathbf{X}$, adjacency matrix $\mathbf{A}$, and node-target matrix $\mathbf{Y}$. \textsc{GCond} distills the given graph by learning a synthetic node-feature matrix $\mathbf{X}_{\mathsf{syn}}$, and using $\mathbf{X}_{\mathsf{syn}}$ to generate $\mathbf{A}_{\mathsf{syn}} \triangleq f_\theta(\mathbf{X}_{\mathsf{syn}})$ which can be realized, \eg, through a parametric similarity function $\operatorname{sim}_\theta(\cdot, \cdot)$ between the features of two nodes, \ie, $\mathbf{A}_{\mathsf{syn}}^{i, j} \triangleq \sigma(\operatorname{sim}_\theta(\mathbf{X}_{\mathsf{syn}}^{i}, \mathbf{X}_{\mathsf{syn}}^{j}))$, where $\sigma(\cdot)$ is the sigmoid function. Finally, both $\mathbf{X}_{\mathsf{syn}}$ and $\theta$ are optimized using the gradient-matching framework proposed by \citet{zhao_dc} (\cref{eqn:gradient_matching}). Another work \citep{graph_distill_arxiv} (GCDM) shares the same framework as \textsc{GCond} but instead uses the distribution matching framework proposed by \citet{dm} (\cref{eqn:distribution_matching}) to optimize $\mathbf{X}_{\mathsf{syn}}$ and $\theta$. Extending to a graph-classification setting, \citet{graph_distill_kdd_22} further propose \textsc{DosCond} with two major changes compared to \textsc{GCond}: (1) instead of parameterizing the adjacency matrix using a similarity function on $\mathbf{X}_{\mathsf{syn}}$, they maintain a free-parameter matrix $\Omega$ with the same size as the adjacency matrix, and sample each $\mathbf{A}_{\mathsf{syn}}^{i, j}$ entry through an independent Bernoulli draw on $\Omega^{i, j}$ as the prior using the reparameterization trick \citep{reparameter}. Such a procedure ensures differentiability as well as discrete matrix synthesis; and (2) $\mathbf{X}_{\mathsf{syn}}$ and $\Omega$ are still optimized using the gradient-matching framework (\cref{eqn:gradient_matching}), albeit with only a single-step, \ie, $T=1$ for improved scalability and without empirically observing a loss in performance.

\paragraph{Recommender Systems.} The amount of online user-feedback data available for training recommender systems is rapidly increasing \citep{data_increasing_recsys}.
% With our ever-increasing online footprint, the amount of logged data of users interacting with smart recommender systems is also rapidly increasing \citep{data_increasing_recsys}. 
Furthermore, typical user-facing recommender systems need to be periodically re-trained \citep{dlrm}, which adds to requirements for smarter data summarization solutions (see \citet{wsdm22} for background on sampling recommender systems data). However, distilling recommender systems data has the following challenges: (1) the data is available in the form of abstract and discrete \texttt{(userID}, \texttt{itemID}, \texttt{relevance)} tuples, which departs from the typical \texttt{(features}, \texttt{label)} setup; (2) the distribution of both user- and item-popularity follows a strong power-law which leads to data scarcity and unstable optimization; and (3) the data inherits a variety of inherent structures, \eg, sequential patterns \citep{svae, sasrec}, user-item graph patterns \citep{gnn_recsys}, item-item co-occurrence patterns \citep{ease}, missing-not-at-randomness \citep{rec_treatments, def_support}, \etc

\citet{inf_ae} propose \textsc{Distill-CF} which distills implicit-feedback recommender systems data, \ie, when the observed user-item relevance is binary (\eg, click or no-click). Such data can be visualized as a binary user-item matrix $\mathbf{R}$ where each row represents a single user, and each column represents an item. On a high-level, \textsc{Distill-CF} synthesizes fake users along with their item-consumption histories, visualized as a synthetic user-item matrix $\mathbf{R}_{\mathsf{syn}}$. Notably, to preserve semantic meaning, the item-space in $\mathbf{R}_{\mathsf{syn}}$ is the same as in $\mathbf{R}$. To alleviate the data discreteness problem, \textsc{Distill-CF} maintains a sampling-prior matrix $\Omega$ which has the same size as $\mathbf{R}_{\mathsf{syn}}$, and can in-turn be used to generate $\mathbf{R}_{\mathsf{syn}}$ using multi-step Gumbel sampling with replacement \citep{gumbel} for each user's prior in $\Omega$ (equivalent to each row). Such a formulation automatically also circumvents the dynamic user- and item-popularity artifact in recommender systems data, which can analogously be controlled by the row- and column-wise entropy of $\Omega$. Finally, $\Omega$ is optimized using the meta-model matching framework proposed by \citet{kip}. Notably, \citet{inf_ae} also propose infinite-width autoencoders which suit the task of item recommendation while also leading to closed-form computation of the inner-loop in the meta-model matching framework (\cref{eqn:kip_optimization}).

\section{Applications} \label{sec:applications}
While the data distillation task was originally designed to accelerate model training, there are numerous other applications of a high-fidelity data summary. Below we briefly discuss a few such promising applications, along with providing pointers to existing works.

\paragraph{Differential Privacy.} Data distillation was recently shown to be a promising solution for differential privacy as defined by \citet{differential_privacy_dwork}. \citet{privacy_free} show that data distillation techniques can perform better than existing state-of-the-art differentially-private data generators \citep{dp_merf, dp_sinkhorn} on both performance and privacy grounds. Notably, the privacy benefits of data distillation techniques are virtually \emph{free}, as none of these methods were optimized for generating differentially-private data. \citet{dd_privacy_clipped} further modify the gradient matching framework (\cref{eqn:gradient_matching}) by clipping and adding white noise to the gradients obtained on the original dataset while optimization. Such a routine was shown to have better sample utility, while also satisfying strict differential privacy guarantees. From a completely application perspective, data distillation has been used to effectively distill sensitive medical data as well \citep{medical_dd_1, medical_dd_2}. \looseness=-1

\paragraph{Neural Architecture Search (NAS).} Automatic searching of neural-network architectures can alleviate a majority of manual effort, as well as lead to more accurate models
% the tedious and heuristic procedure of manually searching neural network architectures 
(see \citet{nas_survey} for a detailed review). Analogous to using model extrapolation, \ie, extrapolating the performance of an under-trained model architecture on the full dataset; data extrapolation, on the other hand, aims to train models on a small, high-fidelity data sample till convergence. 
\citet{zhao_dc} 
% A few data distillation techniques  
show promise of their technique (DC) on a small custom NAS test-bed consisting of only $720$ variations of the ConvNet architecture \citep{conv_net} by employing the data extrapolation framework. However, \citet{dc_bench} show that data distillation \emph{does not} perform well when evaluating diverse architectures on the bigger test-bed, NAS-Bench-201 \citep{nas_201}, calling for better rank-preserving data distillation techniques.

\paragraph{Continual Learning.} Never-ending learning (see \citet{continual} for a detailed review) has been frequently associated with catastrophic forgetting \citep{catast_forgetting}, \ie, patterns extracted from old data/tasks are easily forgotten when patterns from new data/tasks are learned. Data distillation has been shown as an effective solution to alleviate catastrophic forgetting, by simply using the distilled data summary in a replay buffer that is continually updated and used in subsequent data/task training \citep{dd_continual_1, dd_continual_2, dd_continual_3}. \citet{remember_past} show further evidence of a simple \emph{compress-then-recall} strategy outperforming existing state-of-the-art continual learning approaches. Notably, \emph{only} the data summary is stored for each task, and a new model is trained (from scratch) using all previous data summaries, for each new incoming task.

\paragraph{Federated Learning.} Federated or collaborative learning (see \citet{federated_survey} for a detailed survey) involves training a learning algorithm in a decentralized fashion. A standard approach to federated learning is to synchronize local parameter updates to a central server, instead of synchronizing the raw data itself \citep{federated_sync_model}. Data distillation, on the other hand, alleviates the need to synchronize large parametric models across clients and servers, by synchronizing tiny synthesized data summaries to the central server instead. Subsequently, the entire training happens only on the central server. Such data distillation-based federated learning methods \citep{federated_distill_1, federated_distill_2, federated_distill_3, federated_distill_4, federated_distill_5, federated_distill_6} are shown to perform better than model-synchronization based federated learning approaches, while also requiring multiple orders lesser client-server communication.

\section{Challenges \& Future Directions} \label{sec:challenges_future_work}

Despite achieving remarkable progress in data-efficient learning, there are numerous framework-based, theoretical, and application-based directions yet to be explored in data distillation. In what follows, we highlight and discuss such directions for the community to further explore, based either on early evidence or our intuition. \looseness=-1

\paragraph{New data modalities.} Extending the discussion in \cref{sec:data_modalities}, existing data distillation techniques have largely been restricted to cater to image datasets, primarily due to the amenable data-optimization in the continuous pixel-domain of images.
Despite recent efforts in increasing the horizon of data distillation to other data modalities such as graphs \citep{graph_distill_kdd_22, graph_distill_iclr_22} and recommender systems \citep{inf_ae}; each data modality poses its unique challenges and calls for future work, \eg, handling long sequences of time-series data in audio-classification \citep{audio_classification}, video classification \citep{video_classification}, self-driving \citep{waymo}; millions of categorical features in tabular data \citep{dcnv2}; sparse and noisy financial data \citep{stock}; \etc
% developing a unified, principled data distillation framework for inherently sparse and discrete data will be useful for a variety of research communities (\eg, text, tabular-data, extreme classification, \etc).

\paragraph{New predictive tasks.}  Another limitation of existing data distillation techniques is that their underlying optimization is primarily designed for classification scenarios. However, a large number of predictive tasks fail to na\"ively fit into the existing supervised data distillation framework, \eg, image-generation \citep{dalle, stable_diffusion}, language modeling \citep{bert, gpt, llama}, representation learning \citep{byol, simclr}, \etc Further, the aforementioned tasks have gained immense popularity and have seen widespread practical use in the recent years, calling for future work in developing data distillation techniques for more predictive tasks.

\paragraph{Better scaling.} Existing data distillation techniques validate their prowess \emph{only} in the super low-data regime (typically $1-50$ data points per class) due to (i) computational difficulties in synthesizing large data summaries with existing techniques; and (ii) collapse to the random-sampling baseline when synthesizing large data summaries, as noted by \citet{dc_bench}. This calls for future work from both directions --- developing efficient data distillation techniques that are scalable to web-scale datasets, and deeper investigations of the cause and potential fixes of the observed scaling artifacts of existing techniques.

\paragraph{Improved optimization.} A unifying thread across data distillation techniques is an underlying bilevel optimization, which is provably NP-hard even in the linear inner-optimization case \citep{bilevel_np_hard}. Notably, bilevel optimization has been successfully applied in a variety of other applications like meta-learning \citep{maml, metasgd}, hyper-parameter optimization \citep{hyperopt_maclaurin, hyperopt_vicol}, neural architecture search \citep{darts_nas}, coreset construction \citep{bilevel_coresets, bilevel_coresets_bayesian}, \etc Despite its success, many theoretical underpinnings are yet to be explored, \eg, the effect of commonly-used singleton solution assumption \citep{singleton_bilevel}, the effect of over-parameterization on bilevel optimization \citep{bilevel_implicit_bias}, connections to statistical influence functions \citep{influence_functions}, the bias-variance tradeoff \citep{bilevel_bias_variance}, \etc Clearly, an overall better understanding of bilevel optimization will directly enable the development of better data distillation techniques.

\paragraph{Improved data-quality evaluation.} As briefly discussed in \cref{sec:framework}, data synthesized using data distillation is evaluated from performance, efficiency, and transferability standpoints. However, numerous high-stakes use-cases call for being able to train robust models from a variety of angles such as fairness \citep{fair_ml}, adversarial robustness \citep{adv_ml}, \etc Hence, synthesizing data summaries able to support such robust model training is practical and an important direction for future work. Notably, while popular metrics exist for evaluating the robustness of learning algorithms from the aforementioned standpoints, developing such notions at the dataset-level is non-trivial, and with little existing literature \citep{fairness_dpp, adv_sampling}.

\section*{Acknowledgments}
We sincerely thank Zhiwei Deng, Bo Zhao, and George Cazenavette for their feedback on early drafts of this survey.

{\small
\bibliography{references}}

\begin{thebibliography}{129}
\providecommand{\natexlab}[1]{#1}
\providecommand{\url}[1]{\texttt{#1}}
\expandafter\ifx\csname urlstyle\endcsname\relax
  \providecommand{\doi}[1]{doi: #1}\else
  \providecommand{\doi}{doi: \begingroup \urlstyle{rm}\Url}\fi

\bibitem[Abbas et~al.(2023)Abbas, Tirumala, Simig, Ganguli, and
  Morcos]{semdedup}
Amro Abbas, Kushal Tirumala, D{\'a}niel Simig, Surya Ganguli, and Ari~S Morcos.
\newblock Semdedup: Data-efficient learning at web-scale through semantic
  deduplication.
\newblock \emph{arXiv preprint arXiv:2303.09540}, 2023.

\bibitem[Arya et~al.(1998)Arya, Mount, Netanyahu, Silverman, and Wu]{ann}
Sunil Arya, David~M Mount, Nathan~S Netanyahu, Ruth Silverman, and Angela~Y Wu.
\newblock An optimal algorithm for approximate nearest neighbor searching fixed
  dimensions.
\newblock \emph{Journal of the ACM (JACM)}, 45\penalty0 (6):\penalty0 891--923,
  1998.

\bibitem[Ayed \& Hayou(2023)Ayed and Hayou]{data_pruning_limitations}
Fadhel Ayed and Soufiane Hayou.
\newblock Data pruning and neural scaling laws: fundamental limitations of
  score-based algorithms.
\newblock \emph{arXiv preprint arXiv:2302.06960}, 2023.

\bibitem[Bachem et~al.(2017)Bachem, Lucic, and Krause]{coreset_general}
Olivier Bachem, Mario Lucic, and Andreas Krause.
\newblock Practical coreset constructions for machine learning.
\newblock \emph{arXiv preprint arXiv:1703.06476}, 2017.

\bibitem[Bae et~al.(2022)Bae, Ng, Lo, Ghassemi, and
  Grosse]{influence_functions}
Juhan Bae, Nathan Ng, Alston Lo, Marzyeh Ghassemi, and Roger Grosse.
\newblock If influence functions are the answer, then what is the question?
\newblock \emph{arXiv preprint arXiv:2209.05364}, 2022.

\bibitem[Ben-Eliezer \& Yogev(2020)Ben-Eliezer and Yogev]{adv_sampling}
Omri Ben-Eliezer and Eylon Yogev.
\newblock The adversarial robustness of sampling.
\newblock In \emph{Proceedings of the 39th ACM SIGMOD-SIGACT-SIGAI Symposium on
  Principles of Database Systems}, pp.\  49--62, 2020.

\bibitem[Bilmes(2022)]{bilmes_submodularity}
Jeff Bilmes.
\newblock Submodularity in machine learning and artificial intelligence.
\newblock \emph{arXiv preprint arXiv:2202.00132}, 2022.

\bibitem[Bohdal et~al.(2020)Bohdal, Yang, and Hospedales]{label_solve}
Ondrej Bohdal, Yongxin Yang, and Timothy Hospedales.
\newblock Flexible dataset distillation: Learn labels instead of images.
\newblock \emph{arXiv preprint arXiv:2006.08572}, 2020.

\bibitem[Borsos et~al.(2020)Borsos, Mutny, and Krause]{bilevel_coresets}
Zal{\'a}n Borsos, Mojmir Mutny, and Andreas Krause.
\newblock Coresets via bilevel optimization for continual learning and
  streaming.
\newblock \emph{Advances in Neural Information Processing Systems},
  33:\penalty0 14879--14890, 2020.

\bibitem[Brown et~al.(2020)Brown, Mann, Ryder, Subbiah, Kaplan, Dhariwal,
  Neelakantan, Shyam, Sastry, Askell, et~al.]{gpt}
Tom Brown, Benjamin Mann, Nick Ryder, Melanie Subbiah, Jared~D Kaplan, Prafulla
  Dhariwal, Arvind Neelakantan, Pranav Shyam, Girish Sastry, Amanda Askell,
  et~al.
\newblock Language models are few-shot learners.
\newblock \emph{Advances in neural information processing systems},
  33:\penalty0 1877--1901, 2020.

\bibitem[Cao et~al.(2021)Cao, Bie, Vahdat, Fidler, and Kreis]{dp_sinkhorn}
Tianshi Cao, Alex Bie, Arash Vahdat, Sanja Fidler, and Karsten Kreis.
\newblock Don’t generate me: Training differentially private generative
  models with sinkhorn divergence.
\newblock \emph{Advances in Neural Information Processing Systems}, 2021.

\bibitem[Casas et~al.(2020)Casas, Gulino, Liao, and Urtasun]{gnn_self_driving}
Sergio Casas, Cole Gulino, Renjie Liao, and Raquel Urtasun.
\newblock Spagnn: Spatially-aware graph neural networks for relational behavior
  forecasting from sensor data.
\newblock In \emph{2020 IEEE International Conference on Robotics and
  Automation (ICRA)}, pp.\  9491--9497. IEEE, 2020.

\bibitem[Cazenavette et~al.(2022)Cazenavette, Wang, Torralba, Efros, and
  Zhu]{mtt}
George Cazenavette, Tongzhou Wang, Antonio Torralba, Alexei~A Efros, and
  Jun-Yan Zhu.
\newblock Dataset distillation by matching training trajectories.
\newblock In \emph{Proceedings of the IEEE/CVF Conference on Computer Vision
  and Pattern Recognition}, pp.\  4750--4759, 2022.

\bibitem[Celis et~al.(2018)Celis, Keswani, Straszak, Deshpande, Kathuria, and
  Vishnoi]{fairness_dpp}
Elisa Celis, Vijay Keswani, Damian Straszak, Amit Deshpande, Tarun Kathuria,
  and Nisheeth Vishnoi.
\newblock Fair and diverse dpp-based data summarization.
\newblock In \emph{International Conference on Machine Learning}, pp.\
  716--725. PMLR, 2018.

\bibitem[Chen et~al.(2022)Chen, Kerkouche, and Fritz]{dd_privacy_clipped}
Dingfan Chen, Raouf Kerkouche, and Mario Fritz.
\newblock Private set generation with discriminative information.
\newblock In \emph{Proceedings of the Advances in Neural Information Processing
  Systems (NeurIPS)}, 2022.

\bibitem[Chen et~al.(2021)Chen, Zhao, Wang, Li, Liu, Li, Yang, and
  Wang]{graph_billion}
Qi~Chen, Bing Zhao, Haidong Wang, Mingqin Li, Chuanjie Liu, Zengzhong Li, Mao
  Yang, and Jingdong Wang.
\newblock Spann: Highly-efficient billion-scale approximate nearest
  neighborhood search.
\newblock \emph{Advances in Neural Information Processing Systems},
  34:\penalty0 5199--5212, 2021.

\bibitem[Chen et~al.(2020)Chen, Kornblith, Norouzi, and Hinton]{simclr}
Ting Chen, Simon Kornblith, Mohammad Norouzi, and Geoffrey Hinton.
\newblock A simple framework for contrastive learning of visual
  representations.
\newblock In \emph{International conference on machine learning}, pp.\
  1597--1607. PMLR, 2020.

\bibitem[Coleman et~al.(2020)Coleman, Yeh, Mussmann, Mirzasoleiman, Bailis,
  Liang, Leskovec, and Zaharia]{svp}
Cody Coleman, Christopher Yeh, Stephen Mussmann, Baharan Mirzasoleiman, Peter
  Bailis, Percy Liang, Jure Leskovec, and Matei Zaharia.
\newblock Selection via proxy: Efficient data selection for deep learning.
\newblock In \emph{International Conference on Learning Representations}, 2020.

\bibitem[Cui et~al.(2022{\natexlab{a}})Cui, Wang, Si, and Hsieh]{dc_bench}
Justin Cui, Ruochen Wang, Si~Si, and Cho-Jui Hsieh.
\newblock {DC}-{BENCH}: Dataset condensation benchmark.
\newblock In \emph{Thirty-sixth Conference on Neural Information Processing
  Systems Datasets and Benchmarks Track}, 2022{\natexlab{a}}.

\bibitem[Cui et~al.(2022{\natexlab{b}})Cui, Wang, Si, and Hsieh]{tesla}
Justin Cui, Ruochen Wang, Si~Si, and Cho-Jui Hsieh.
\newblock Scaling up dataset distillation to imagenet-1k with constant memory.
\newblock \emph{arXiv preprint arXiv:2211.10586}, 2022{\natexlab{b}}.

\bibitem[Deng \& Russakovsky(2022)Deng and Russakovsky]{remember_past}
Zhiwei Deng and Olga Russakovsky.
\newblock Remember the past: Distilling datasets into addressable memories for
  neural networks.
\newblock In \emph{Advances in Neural Information Processing Systems}, 2022.

\bibitem[Devlin et~al.(2019)Devlin, Chang, Lee, and Toutanova]{bert}
Jacob Devlin, Ming-Wei Chang, Kenton Lee, and Kristina Toutanova.
\newblock {BERT}: Pre-training of deep bidirectional transformers for language
  understanding.
\newblock In \emph{Proceedings of the 2019 Conference of the North {A}merican
  Chapter of the Association for Computational Linguistics: Human Language
  Technologies, Volume 1 (Long and Short Papers)}, Minneapolis, Minnesota, June
  2019. Association for Computational Linguistics.

\bibitem[Dong et~al.(2016)Dong, Loy, and Tang]{fsrcnn}
Chao Dong, Chen~Change Loy, and Xiaoou Tang.
\newblock Accelerating the super-resolution convolutional neural network.
\newblock In \emph{European conference on computer vision}, pp.\  391--407.
  Springer, 2016.

\bibitem[Dong et~al.(2022)Dong, Zhao, and Lyu]{privacy_free}
Tian Dong, Bo~Zhao, and Lingjuan Lyu.
\newblock Privacy for free: How does dataset condensation help privacy?
\newblock In \emph{Proceedings of the 39th International Conference on Machine
  Learning}. PMLR, 2022.

\bibitem[Dong \& Yang(2020)Dong and Yang]{nas_201}
Xuanyi Dong and Yi~Yang.
\newblock Nas-bench-201: Extending the scope of reproducible neural
  architecture search.
\newblock In \emph{International Conference on Learning Representations}, 2020.

\bibitem[Dwork(2008)]{differential_privacy_dwork}
Cynthia Dwork.
\newblock Differential privacy: A survey of results.
\newblock In \emph{International conference on theory and applications of
  models of computation}, pp.\  1--19. Springer, 2008.

\bibitem[Elsken et~al.(2019)Elsken, Metzen, and Hutter]{nas_survey}
Thomas Elsken, Jan~Hendrik Metzen, and Frank Hutter.
\newblock Neural architecture search: A survey.
\newblock \emph{The Journal of Machine Learning Research}, 20\penalty0
  (1):\penalty0 1997--2017, 2019.

\bibitem[Fan et~al.(2019)Fan, Ma, Li, He, Zhao, Tang, and Yin]{gnn_social}
Wenqi Fan, Yao Ma, Qing Li, Yuan He, Eric Zhao, Jiliang Tang, and Dawei Yin.
\newblock Graph neural networks for social recommendation.
\newblock In \emph{The world wide web conference}, pp.\  417--426, 2019.

\bibitem[Finn et~al.(2017)Finn, Abbeel, and Levine]{maml}
Chelsea Finn, Pieter Abbeel, and Sergey Levine.
\newblock Model-agnostic meta-learning for fast adaptation of deep networks.
\newblock In \emph{International conference on machine learning}, pp.\
  1126--1135. PMLR, 2017.

\bibitem[Franceschi et~al.(2018)Franceschi, Frasconi, Salzo, Grazzi, and
  Pontil]{singleton_bilevel}
Luca Franceschi, Paolo Frasconi, Saverio Salzo, Riccardo Grazzi, and
  Massimiliano Pontil.
\newblock Bilevel programming for hyperparameter optimization and
  meta-learning.
\newblock In \emph{International Conference on Machine Learning}, pp.\
  1568--1577. PMLR, 2018.

\bibitem[French(1999)]{catast_forgetting}
Robert~M French.
\newblock Catastrophic forgetting in connectionist networks.
\newblock \emph{Trends in cognitive sciences}, 3\penalty0 (4):\penalty0
  128--135, 1999.

\bibitem[Geiping \& Goldstein(2022)Geiping and Goldstein]{cramming}
Jonas Geiping and Tom Goldstein.
\newblock Cramming: Training a language model on a single gpu in one day, 2022.

\bibitem[Ghorbani \& Zou(2019)Ghorbani and Zou]{data_shapley}
Amirata Ghorbani and James Zou.
\newblock Data shapley: Equitable valuation of data for machine learning.
\newblock In \emph{International Conference on Machine Learning}, pp.\
  2242--2251. PMLR, 2019.

\bibitem[Ghorbani et~al.(2021)Ghorbani, Firat, Freitag, Bapna, Krikun, Garcia,
  Chelba, and Cherry]{scaling_2}
Behrooz Ghorbani, Orhan Firat, Markus Freitag, Ankur Bapna, Maxim Krikun,
  Xavier Garcia, Ciprian Chelba, and Colin Cherry.
\newblock Scaling laws for neural machine translation.
\newblock \emph{arXiv preprint arXiv:2109.07740}, 2021.

\bibitem[Gidaris \& Komodakis(2018)Gidaris and Komodakis]{conv_net}
Spyros Gidaris and Nikos Komodakis.
\newblock Dynamic few-shot visual learning without forgetting.
\newblock In \emph{Proceedings of the IEEE conference on computer vision and
  pattern recognition}, pp.\  4367--4375, 2018.

\bibitem[Goetz \& Tewari(2020)Goetz and Tewari]{federated_distill_1}
Jack Goetz and Ambuj Tewari.
\newblock Federated learning via synthetic data.
\newblock \emph{arXiv preprint arXiv:2008.04489}, 2020.

\bibitem[Goodfellow et~al.(2014)Goodfellow, Pouget-Abadie, Mirza, Xu,
  Warde-Farley, Ozair, Courville, and Bengio]{gan}
Ian Goodfellow, Jean Pouget-Abadie, Mehdi Mirza, Bing Xu, David Warde-Farley,
  Sherjil Ozair, Aaron Courville, and Yoshua Bengio.
\newblock Generative adversarial nets.
\newblock In \emph{Advances in neural information processing systems}, pp.\
  2672--2680, 2014.

\bibitem[Grill et~al.(2020)Grill, Strub, Altch{\'e}, Tallec, Richemond,
  Buchatskaya, Doersch, Avila~Pires, Guo, Gheshlaghi~Azar, et~al.]{byol}
Jean-Bastien Grill, Florian Strub, Florent Altch{\'e}, Corentin Tallec, Pierre
  Richemond, Elena Buchatskaya, Carl Doersch, Bernardo Avila~Pires, Zhaohan
  Guo, Mohammad Gheshlaghi~Azar, et~al.
\newblock Bootstrap your own latent-a new approach to self-supervised learning.
\newblock \emph{Advances in neural information processing systems},
  33:\penalty0 21271--21284, 2020.

\bibitem[Guo et~al.(2022)Guo, Zhao, and Bai]{deepcore}
Chengcheng Guo, Bo~Zhao, and Yanbing Bai.
\newblock Deepcore: A comprehensive library for coreset selection in deep
  learning.
\newblock In \emph{Database and Expert Systems Applications: 33rd International
  Conference, DEXA 2022, Vienna, Austria, August 22--24, 2022, Proceedings,
  Part I}, pp.\  181--195. Springer, 2022.

\bibitem[Gupta et~al.(2022)Gupta, Kim, Lee, Tse, Lee, Wei, Brooks, and
  Wu]{chasing_carbon}
Udit Gupta, Young~Geun Kim, Sylvia Lee, Jordan Tse, Hsien-Hsin~S Lee, Gu-Yeon
  Wei, David Brooks, and Carole-Jean Wu.
\newblock Chasing carbon: The elusive environmental footprint of computing.
\newblock \emph{IEEE Micro}, 42\penalty0 (4):\penalty0 37--47, 2022.

\bibitem[Harder et~al.(2021)Harder, Adamczewski, and Park]{dp_merf}
Frederik Harder, Kamil Adamczewski, and Mijung Park.
\newblock Dp-merf: Differentially private mean embeddings with randomfeatures
  for practical privacy-preserving data generation.
\newblock In \emph{International conference on artificial intelligence and
  statistics}, pp.\  1819--1827. PMLR, 2021.

\bibitem[Hershey et~al.(2017)Hershey, Chaudhuri, Ellis, Gemmeke, Jansen, Moore,
  Plakal, Platt, Saurous, Seybold, et~al.]{audio_classification}
Shawn Hershey, Sourish Chaudhuri, Daniel~PW Ellis, Jort~F Gemmeke, Aren Jansen,
  R~Channing Moore, Manoj Plakal, Devin Platt, Rif~A Saurous, Bryan Seybold,
  et~al.
\newblock Cnn architectures for large-scale audio classification.
\newblock In \emph{2017 ieee international conference on acoustics, speech and
  signal processing (icassp)}, pp.\  131--135. IEEE, 2017.

\bibitem[Hinton et~al.(2015)Hinton, Vinyals, Dean,
  et~al.]{knowledge_distillation}
Geoffrey Hinton, Oriol Vinyals, Jeff Dean, et~al.
\newblock Distilling the knowledge in a neural network.
\newblock \emph{arXiv preprint arXiv:1503.02531}, 2\penalty0 (7), 2015.

\bibitem[Hoffmann et~al.(2022)Hoffmann, Borgeaud, Mensch, Buchatskaya, Cai,
  Rutherford, Casas, Hendricks, Welbl, Clark, et~al.]{scaling_3}
Jordan Hoffmann, Sebastian Borgeaud, Arthur Mensch, Elena Buchatskaya, Trevor
  Cai, Eliza Rutherford, Diego de~Las Casas, Lisa~Anne Hendricks, Johannes
  Welbl, Aidan Clark, et~al.
\newblock Training compute-optimal large language models.
\newblock \emph{arXiv preprint arXiv:2203.15556}, 2022.

\bibitem[Hu et~al.(2022)Hu, Goetz, Malik, Zhan, Liu, and
  Liu]{federated_distill_3}
Shengyuan Hu, Jack Goetz, Kshitiz Malik, Hongyuan Zhan, Zhe Liu, and Yue Liu.
\newblock Fedsynth: Gradient compression via synthetic data in federated
  learning.
\newblock \emph{arXiv preprint arXiv:2204.01273}, 2022.

\bibitem[Jacot et~al.(2018)Jacot, Gabriel, and Hongler]{ntk}
Arthur Jacot, Franck Gabriel, and Cl{\'e}ment Hongler.
\newblock Neural tangent kernel: Convergence and generalization in neural
  networks.
\newblock \emph{Advances in neural information processing systems}, 31, 2018.

\bibitem[Jang et~al.(2017)Jang, Gu, and Poole]{gumbel}
Eric Jang, Shixiang Gu, and Ben Poole.
\newblock Categorical reparameterization with gumbel-softmax.
\newblock In \emph{5th International Conference on Learning Representations,
  {ICLR} 2017, Toulon, France, April 24-26, 2017, Conference Track
  Proceedings}. OpenReview.net, 2017.

\bibitem[Jin et~al.(2022{\natexlab{a}})Jin, Tang, Jiang, Li, Zhang, Tang, and
  Yin]{graph_distill_kdd_22}
Wei Jin, Xianfeng Tang, Haoming Jiang, Zheng Li, Danqing Zhang, Jiliang Tang,
  and Bing Yin.
\newblock Condensing graphs via one-step gradient matching.
\newblock In \emph{Proceedings of the 28th ACM SIGKDD Conference on Knowledge
  Discovery and Data Mining}, pp.\  720--730, 2022{\natexlab{a}}.

\bibitem[Jin et~al.(2022{\natexlab{b}})Jin, Zhao, Zhang, Liu, Tang, and
  Shah]{graph_distill_iclr_22}
Wei Jin, Lingxiao Zhao, Shichang Zhang, Yozen Liu, Jiliang Tang, and Neil Shah.
\newblock Graph condensation for graph neural networks.
\newblock In \emph{International Conference on Learning Representations},
  2022{\natexlab{b}}.

\bibitem[Kang \& McAuley(2018)Kang and McAuley]{sasrec}
Wang-Cheng Kang and Julian McAuley.
\newblock Self-attentive sequential recommendation.
\newblock In \emph{2018 IEEE international conference on data mining (ICDM)},
  pp.\  197--206. IEEE, 2018.

\bibitem[Kaplan et~al.(2020)Kaplan, McCandlish, Henighan, Brown, Chess, Child,
  Gray, Radford, Wu, and Amodei]{scaling_1}
Jared Kaplan, Sam McCandlish, Tom Henighan, Tom~B Brown, Benjamin Chess, Rewon
  Child, Scott Gray, Alec Radford, Jeffrey Wu, and Dario Amodei.
\newblock Scaling laws for neural language models.
\newblock \emph{arXiv preprint arXiv:2001.08361}, 2020.

\bibitem[Karpathy et~al.(2014)Karpathy, Toderici, Shetty, Leung, Sukthankar,
  and Fei-Fei]{video_classification}
Andrej Karpathy, George Toderici, Sanketh Shetty, Thomas Leung, Rahul
  Sukthankar, and Li~Fei-Fei.
\newblock Large-scale video classification with convolutional neural networks.
\newblock In \emph{Proceedings of the IEEE conference on Computer Vision and
  Pattern Recognition}, pp.\  1725--1732, 2014.

\bibitem[Killamsetty et~al.(2021)Killamsetty, S, Ramakrishnan, De, and
  Iyer]{grad_match}
Krishnateja Killamsetty, Durga S, Ganesh Ramakrishnan, Abir De, and Rishabh
  Iyer.
\newblock Grad-match: Gradient matching based data subset selection for
  efficient deep model training.
\newblock In Marina Meila and Tong Zhang (eds.), \emph{Proceedings of the 38th
  International Conference on Machine Learning}, volume 139 of
  \emph{Proceedings of Machine Learning Research}, pp.\  5464--5474. PMLR,
  18--24 Jul 2021.

\bibitem[Kim et~al.(2022)Kim, Kim, Oh, Yun, Song, Jeong, Ha, and Song]{idc}
Jang-Hyun Kim, Jinuk Kim, Seong~Joon Oh, Sangdoo Yun, Hwanjun Song, Joonhyun
  Jeong, Jung-Woo Ha, and Hyun~Oh Song.
\newblock Dataset condensation via efficient synthetic-data parameterization.
\newblock In \emph{Proceedings of the 39th International Conference on Machine
  Learning}, 2022.

\bibitem[Kipf \& Welling(2017)Kipf and Welling]{gcn}
Thomas~N. Kipf and Max Welling.
\newblock {Semi-Supervised Classification with Graph Convolutional Networks}.
\newblock In \emph{Proceedings of the 5th International Conference on Learning
  Representations}, ICLR '17, 2017.

\bibitem[Kone{\v{c}}n{\`y} et~al.(2016)Kone{\v{c}}n{\`y}, McMahan, Ramage, and
  Richt{\'a}rik]{federated_sync_model}
Jakub Kone{\v{c}}n{\`y}, H~Brendan McMahan, Daniel Ramage, and Peter
  Richt{\'a}rik.
\newblock Federated optimization: Distributed machine learning for on-device
  intelligence.
\newblock \emph{arXiv preprint arXiv:1610.02527}, 2016.

\bibitem[LeCun et~al.(1989)LeCun, Denker, and Solla]{model_compression}
Yann LeCun, John Denker, and Sara Solla.
\newblock Optimal brain damage.
\newblock \emph{Advances in neural information processing systems}, 2, 1989.

\bibitem[Lee et~al.(2022{\natexlab{a}})Lee, Lee, and Hwang]{kfs}
Hae~Beom Lee, Dong~Bok Lee, and Sung~Ju Hwang.
\newblock Dataset condensation with latent space knowledge factorization and
  sharing.
\newblock \emph{arXiv preprint arXiv:2208.10494}, 2022{\natexlab{a}}.

\bibitem[Lee et~al.(2020)Lee, Schoenholz, Pennington, Adlam, Xiao, Novak, and
  Sohl-Dickstein]{finite_vs_infinite_2}
Jaehoon Lee, Samuel Schoenholz, Jeffrey Pennington, Ben Adlam, Lechao Xiao,
  Roman Novak, and Jascha Sohl-Dickstein.
\newblock Finite versus infinite neural networks: an empirical study.
\newblock \emph{Advances in Neural Information Processing Systems},
  33:\penalty0 15156--15172, 2020.

\bibitem[Lee et~al.(2022{\natexlab{b}})Lee, Chun, Jung, Yun, and Yoon]{dcc}
Saehyung Lee, Sanghyuk Chun, Sangwon Jung, Sangdoo Yun, and Sungroh Yoon.
\newblock Dataset condensation with contrastive signals.
\newblock In \emph{Proceedings of the 39th International Conference on Machine
  Learning}, pp.\  12352--12364, 2022{\natexlab{b}}.

\bibitem[Li et~al.(2020{\natexlab{a}})Li, Togo, Ogawa, and
  Haseyama]{medical_dd_1}
Guang Li, Ren Togo, Takahiro Ogawa, and Miki Haseyama.
\newblock Soft-label anonymous gastric x-ray image distillation.
\newblock In \emph{2020 IEEE International Conference on Image Processing
  (ICIP)}, pp.\  305--309. IEEE, 2020{\natexlab{a}}.

\bibitem[Li et~al.(2022)Li, Togo, Ogawa, and Haseyama]{medical_dd_2}
Guang Li, Ren Togo, Takahiro Ogawa, and Miki Haseyama.
\newblock Compressed gastric image generation based on soft-label dataset
  distillation for medical data sharing.
\newblock \emph{Computer Methods and Programs in Biomedicine}, pp.\  107189,
  2022.

\bibitem[Li et~al.(2020{\natexlab{b}})Li, Sahu, Talwalkar, and
  Smith]{federated_survey}
Tian Li, Anit~Kumar Sahu, Ameet Talwalkar, and Virginia Smith.
\newblock Federated learning: Challenges, methods, and future directions.
\newblock \emph{IEEE Signal Processing Magazine}, 37\penalty0 (3):\penalty0
  50--60, 2020{\natexlab{b}}.

\bibitem[Li et~al.(2017)Li, Zhou, Chen, and Li]{metasgd}
Zhenguo Li, Fengwei Zhou, Fei Chen, and Hang Li.
\newblock Meta-sgd: Learning to learn quickly for few-shot learning.
\newblock \emph{arXiv preprint arXiv:1707.09835}, 2017.

\bibitem[Liu et~al.(2019)Liu, Simonyan, and Yang]{darts_nas}
Hanxiao Liu, Karen Simonyan, and Yiming Yang.
\newblock {DARTS}: Differentiable architecture search.
\newblock In \emph{International Conference on Learning Representations}, 2019.

\bibitem[Liu et~al.(2022{\natexlab{a}})Liu, Li, Chen, and
  Song]{graph_distill_arxiv}
Mengyang Liu, Shanchuan Li, Xinshi Chen, and Le~Song.
\newblock Graph condensation via receptive field distribution matching.
\newblock \emph{arXiv preprint arXiv:2206.13697}, 2022{\natexlab{a}}.

\bibitem[Liu et~al.(2022{\natexlab{b}})Liu, Yu, and Zhou]{federated_distill_6}
Ping Liu, Xin Yu, and Joey~Tianyi Zhou.
\newblock Meta knowledge condensation for federated learning.
\newblock \emph{arXiv preprint arXiv:2209.14851}, 2022{\natexlab{b}}.

\bibitem[Liu et~al.(2022{\natexlab{c}})Liu, Wang, Yang, Ye, and Wang]{haba}
Songhua Liu, Kai Wang, Xingyi Yang, Jingwen Ye, and Xinchao Wang.
\newblock Dataset distillation via factorization.
\newblock \emph{NeurIPS}, 2022{\natexlab{c}}.

\bibitem[Loo et~al.(2022)Loo, Hasani, Amini, and Rus]{rfad}
Noel Loo, Ramin Hasani, Alexander Amini, and Daniela Rus.
\newblock Efficient dataset distillation using random feature approximation.
\newblock In \emph{Advances in Neural Information Processing Systems}, 2022.

\bibitem[Lorraine et~al.(2020)Lorraine, Vicol, and Duvenaud]{hyperopt_vicol}
Jonathan Lorraine, Paul Vicol, and David Duvenaud.
\newblock Optimizing millions of hyperparameters by implicit differentiation.
\newblock In \emph{International Conference on Artificial Intelligence and
  Statistics}, pp.\  1540--1552. PMLR, 2020.

\bibitem[Maclaurin et~al.(2015)Maclaurin, Duvenaud, and
  Adams]{hyperopt_maclaurin}
Dougal Maclaurin, David Duvenaud, and Ryan Adams.
\newblock Gradient-based hyperparameter optimization through reversible
  learning.
\newblock In \emph{International conference on machine learning}, pp.\
  2113--2122. PMLR, 2015.

\bibitem[Maddison et~al.(2017)Maddison, Mnih, and Teh]{reparameter}
Chris~J. Maddison, Andriy Mnih, and Yee~Whye Teh.
\newblock The concrete distribution: A continuous relaxation of discrete random
  variables.
\newblock In \emph{International Conference on Learning Representations}, 2017.

\bibitem[Madry et~al.(2018)Madry, Makelov, Schmidt, Tsipras, and Vladu]{adv_ml}
Aleksander Madry, Aleksandar Makelov, Ludwig Schmidt, Dimitris Tsipras, and
  Adrian Vladu.
\newblock Towards deep learning models resistant to adversarial attacks.
\newblock In \emph{International Conference on Learning Representations}, 2018.

\bibitem[McLachlan \& Krishnan(2007)McLachlan and Krishnan]{em}
Geoffrey~J McLachlan and Thriyambakam Krishnan.
\newblock \emph{The EM algorithm and extensions}.
\newblock John Wiley \& Sons, 2007.

\bibitem[Mehrabi et~al.(2021)Mehrabi, Morstatter, Saxena, Lerman, and
  Galstyan]{fair_ml}
Ninareh Mehrabi, Fred Morstatter, Nripsuta Saxena, Kristina Lerman, and Aram
  Galstyan.
\newblock A survey on bias and fairness in machine learning.
\newblock \emph{ACM Computing Surveys (CSUR)}, 54\penalty0 (6):\penalty0 1--35,
  2021.

\bibitem[Metz et~al.(2019)Metz, Maheswaranathan, Nixon, Freeman, and
  Sohl-Dickstein]{bptt_loss_landscape}
Luke Metz, Niru Maheswaranathan, Jeremy Nixon, Daniel Freeman, and Jascha
  Sohl-Dickstein.
\newblock Understanding and correcting pathologies in the training of learned
  optimizers.
\newblock In \emph{International Conference on Machine Learning}, pp.\
  4556--4565. PMLR, 2019.

\bibitem[Mirzasoleiman et~al.(2020)Mirzasoleiman, Bilmes, and Leskovec]{craig}
Baharan Mirzasoleiman, Jeff Bilmes, and Jure Leskovec.
\newblock Coresets for data-efficient training of machine learning models.
\newblock In \emph{International Conference on Machine Learning}, pp.\
  6950--6960. PMLR, 2020.

\bibitem[Mittal et~al.(2021)Mittal, Sachdeva, Agrawal, Agarwal, Kar, and
  Varma]{eclare}
Anshul Mittal, Noveen Sachdeva, Sheshansh Agrawal, Sumeet Agarwal, Purushottam
  Kar, and Manik Varma.
\newblock Eclare: Extreme classification with label graph correlations.
\newblock In \emph{Proceedings of the Web Conference 2021}, WWW '21, 2021.

\bibitem[Naumov et~al.(2019)Naumov, Mudigere, Shi, Huang, Sundaraman, Park,
  Wang, Gupta, Wu, Azzolini, Dzhulgakov, Mallevich, Cherniavskii, Lu,
  Krishnamoorthi, Yu, Kondratenko, Pereira, Chen, Chen, Rao, Jia, Xiong, and
  Smelyanskiy]{dlrm}
Maxim Naumov, Dheevatsa Mudigere, Hao-Jun~Michael Shi, Jianyu Huang, Narayanan
  Sundaraman, Jongsoo Park, Xiaodong Wang, Udit Gupta, Carole-Jean Wu,
  Alisson~G. Azzolini, Dmytro Dzhulgakov, Andrey Mallevich, Ilia Cherniavskii,
  Yinghai Lu, Raghuraman Krishnamoorthi, Ansha Yu, Volodymyr Kondratenko,
  Stephanie Pereira, Xianjie Chen, Wenlin Chen, Vijay Rao, Bill Jia, Liang
  Xiong, and Misha Smelyanskiy.
\newblock Deep learning recommendation model for personalization and
  recommendation systems.
\newblock \emph{CoRR}, abs/1906.00091, 2019.

\bibitem[Neal(2012)]{nngp}
Radford~M Neal.
\newblock \emph{Bayesian learning for neural networks}, volume 118.
\newblock Springer Science \& Business Media, 2012.

\bibitem[Nguyen et~al.(2021{\natexlab{a}})Nguyen, Chen, and Lee]{kip}
Timothy Nguyen, Zhourong Chen, and Jaehoon Lee.
\newblock Dataset meta-learning from kernel ridge-regression.
\newblock In \emph{International Conference on Learning Representations},
  2021{\natexlab{a}}.

\bibitem[Nguyen et~al.(2021{\natexlab{b}})Nguyen, Novak, Xiao, and
  Lee]{kip_conv}
Timothy Nguyen, Roman Novak, Lechao Xiao, and Jaehoon Lee.
\newblock Dataset distillation with infinitely wide convolutional networks.
\newblock \emph{Advances in Neural Information Processing Systems}, 34,
  2021{\natexlab{b}}.

\bibitem[Parisi et~al.(2019)Parisi, Kemker, Part, Kanan, and
  Wermter]{continual}
German~I Parisi, Ronald Kemker, Jose~L Part, Christopher Kanan, and Stefan
  Wermter.
\newblock Continual lifelong learning with neural networks: A review.
\newblock \emph{Neural Networks}, 113:\penalty0 54--71, 2019.

\bibitem[Pratt(1992)]{transfer_learning}
Lorien~Y Pratt.
\newblock Discriminability-based transfer between neural networks.
\newblock \emph{Advances in neural information processing systems}, 5, 1992.

\bibitem[Ramesh et~al.(2022)Ramesh, Dhariwal, Nichol, Chu, and Chen]{dalle}
Aditya Ramesh, Prafulla Dhariwal, Alex Nichol, Casey Chu, and Mark Chen.
\newblock Hierarchical text-conditional image generation with clip latents.
\newblock \emph{arXiv preprint arXiv:2204.06125}, 2022.

\bibitem[Rombach et~al.(2022)Rombach, Blattmann, Lorenz, Esser, and
  Ommer]{stable_diffusion}
Robin Rombach, Andreas Blattmann, Dominik Lorenz, Patrick Esser, and Bj{\"o}rn
  Ommer.
\newblock High-resolution image synthesis with latent diffusion models.
\newblock In \emph{Proceedings of the IEEE/CVF Conference on Computer Vision
  and Pattern Recognition}, pp.\  10684--10695, 2022.

\bibitem[Rosasco et~al.(2021)Rosasco, Carta, Cossu, Lomonaco, and
  Bacciu]{dd_continual_2}
Andrea Rosasco, Antonio Carta, Andrea Cossu, Vincenzo Lomonaco, and Davide
  Bacciu.
\newblock Distilled replay: Overcoming forgetting through synthetic samples.
\newblock \emph{arXiv preprint arXiv:2103.15851}, 2021.

\bibitem[S et~al.(2021)S, Iyer, Ramakrishnan, and De]{selcon_coreset}
Durga S, Rishabh Iyer, Ganesh Ramakrishnan, and Abir De.
\newblock Training data subset selection for regression with controlled
  generalization error.
\newblock In Marina Meila and Tong Zhang (eds.), \emph{Proceedings of the 38th
  International Conference on Machine Learning}, volume 139 of
  \emph{Proceedings of Machine Learning Research}, pp.\  9202--9212. PMLR,
  18--24 Jul 2021.

\bibitem[Sachdeva \& McAuley(2020)Sachdeva and McAuley]{reviews_sigir}
Noveen Sachdeva and Julian McAuley.
\newblock \emph{How Useful Are Reviews for Recommendation? A Critical Review
  and Potential Improvements}, pp.\  1845–1848.
\newblock SIGIR '20. Association for Computing Machinery, New York, NY, USA,
  2020.
\newblock ISBN 9781450380164.
\newblock \doi{10.1145/3397271.3401281}.

\bibitem[Sachdeva et~al.(2019)Sachdeva, Manco, Ritacco, and Pudi]{svae}
Noveen Sachdeva, Giuseppe Manco, Ettore Ritacco, and Vikram Pudi.
\newblock Sequential variational autoencoders for collaborative filtering.
\newblock In \emph{Proceedings of the twelfth ACM international conference on
  web search and data mining}, pp.\  600--608, 2019.

\bibitem[Sachdeva et~al.(2020)Sachdeva, Su, and Joachims]{def_support}
Noveen Sachdeva, Yi~Su, and Thorsten Joachims.
\newblock Off-policy bandits with deficient support.
\newblock In \emph{Proceedings of the 26th ACM SIGKDD International Conference
  on Knowledge Discovery \& Data Mining}, KDD '20, pp.\  965–975, New York,
  NY, USA, 2020. Association for Computing Machinery.
\newblock \doi{10.1145/3394486.3403139}.

\bibitem[Sachdeva et~al.(2022{\natexlab{a}})Sachdeva, Dhaliwal, Wu, and
  McAuley]{inf_ae}
Noveen Sachdeva, Mehak~Preet Dhaliwal, Carole-Jean Wu, and Julian McAuley.
\newblock Infinite recommendation networks: A data-centric approach.
\newblock In \emph{Advances in Neural Information Processing Systems},
  2022{\natexlab{a}}.

\bibitem[Sachdeva et~al.(2022{\natexlab{b}})Sachdeva, Wang, Han, Gupta, and
  McAuley]{gapformer}
Noveen Sachdeva, Ziran Wang, Kyungtae Han, Rohit Gupta, and Julian McAuley.
\newblock Gapformer: Fast autoregressive transformers meet rnns for
  personalized adaptive cruise control.
\newblock In \emph{2022 IEEE 25th International Conference on Intelligent
  Transportation Systems (ITSC)}, pp.\  2528--2535, 2022{\natexlab{b}}.
\newblock \doi{10.1109/ITSC55140.2022.9922275}.

\bibitem[Sachdeva et~al.(2022{\natexlab{c}})Sachdeva, Wu, and McAuley]{wsdm22}
Noveen Sachdeva, Carole-Jean Wu, and Julian McAuley.
\newblock On sampling collaborative filtering datasets.
\newblock In \emph{Proceedings of the Fifteenth ACM International Conference on
  Web Search and Data Mining}, WSDM '22, 2022{\natexlab{c}}.

\bibitem[Sangermano et~al.(2022)Sangermano, Carta, Cossu, and
  Bacciu]{dd_continual_3}
Mattia Sangermano, Antonio Carta, Andrea Cossu, and Davide Bacciu.
\newblock Sample condensation in online continual learning.
\newblock In \emph{2022 International Joint Conference on Neural Networks
  (IJCNN)}, pp.\  01--08. IEEE, 2022.

\bibitem[Scao et~al.(2022)Scao, Fan, Akiki, Pavlick, Ili{\'c}, Hesslow,
  Castagn{\'e}, Luccioni, Yvon, Gall{\'e}, et~al.]{bloom}
Teven~Le Scao, Angela Fan, Christopher Akiki, Ellie Pavlick, Suzana Ili{\'c},
  Daniel Hesslow, Roman Castagn{\'e}, Alexandra~Sasha Luccioni, Fran{\c{c}}ois
  Yvon, Matthias Gall{\'e}, et~al.
\newblock Bloom: A 176b-parameter open-access multilingual language model.
\newblock \emph{arXiv preprint arXiv:2211.05100}, 2022.

\bibitem[Schirrmeister et~al.(2022)Schirrmeister, Liu, Hooker, and
  Ball]{less_is_more}
Robin~Tibor Schirrmeister, Rosanne Liu, Sara Hooker, and Tonio Ball.
\newblock When less is more: Simplifying inputs aids neural network
  understanding.
\newblock \emph{arXiv preprint arXiv:2201.05610}, 2022.

\bibitem[Schnabel et~al.(2016)Schnabel, Swaminathan, Singh, Chandak, and
  Joachims]{rec_treatments}
Tobias Schnabel, Adith Swaminathan, Ashudeep Singh, Navin Chandak, and Thorsten
  Joachims.
\newblock Recommendations as treatments: Debiasing learning and evaluation.
\newblock In \emph{Proceedings of The 33rd International Conference on Machine
  Learning}, volume~48 of \emph{Proceedings of Machine Learning Research}, pp.\
   1670--1679. PMLR, 2016.

\bibitem[Song et~al.(2022)Song, Liu, Chen, Festag, Trinitis, Schulz, and
  Knoll]{federated_distill_5}
Rui Song, Dai Liu, Dave~Zhenyu Chen, Andreas Festag, Carsten Trinitis, Martin
  Schulz, and Alois Knoll.
\newblock Federated learning via decentralized dataset distillation in
  resource-constrained edge environments.
\newblock \emph{arXiv preprint arXiv:2208.11311}, 2022.

\bibitem[Sorscher et~al.(2022)Sorscher, Geirhos, Shekhar, Ganguli, and
  Morcos]{data_quality}
Ben Sorscher, Robert Geirhos, Shashank Shekhar, Surya Ganguli, and Ari~S.
  Morcos.
\newblock Beyond neural scaling laws: beating power law scaling via data
  pruning.
\newblock In Alice~H. Oh, Alekh Agarwal, Danielle Belgrave, and Kyunghyun Cho
  (eds.), \emph{Advances in Neural Information Processing Systems}, 2022.

\bibitem[Steck(2019)]{ease}
Harald Steck.
\newblock Embarrassingly shallow autoencoders for sparse data.
\newblock In \emph{The World Wide Web Conference}, 2019.

\bibitem[Sucholutsky \& Schonlau(2021)Sucholutsky and Schonlau]{text_distill}
Ilia Sucholutsky and Matthias Schonlau.
\newblock Soft-label dataset distillation and text dataset distillation.
\newblock In \emph{2021 International Joint Conference on Neural Networks
  (IJCNN)}, pp.\  1--8. IEEE, 2021.

\bibitem[Sun et~al.(2020)Sun, Kretzschmar, Dotiwalla, Chouard, Patnaik, Tsui,
  Guo, Zhou, Chai, Caine, et~al.]{waymo}
Pei Sun, Henrik Kretzschmar, Xerxes Dotiwalla, Aurelien Chouard, Vijaysai
  Patnaik, Paul Tsui, James Guo, Yin Zhou, Yuning Chai, Benjamin Caine, et~al.
\newblock Scalability in perception for autonomous driving: Waymo open dataset.
\newblock In \emph{Proceedings of the IEEE/CVF conference on computer vision
  and pattern recognition}, pp.\  2446--2454, 2020.

\bibitem[Thoppilan et~al.(2022)Thoppilan, De~Freitas, Hall, Shazeer,
  Kulshreshtha, Cheng, Jin, Bos, Baker, Du, et~al.]{lamda}
Romal Thoppilan, Daniel De~Freitas, Jamie Hall, Noam Shazeer, Apoorv
  Kulshreshtha, Heng-Tze Cheng, Alicia Jin, Taylor Bos, Leslie Baker, Yu~Du,
  et~al.
\newblock Lamda: Language models for dialog applications.
\newblock \emph{arXiv preprint arXiv:2201.08239}, 2022.

\bibitem[Toneva et~al.(2019)Toneva, Sordoni, des Combes, Trischler, Bengio, and
  Gordon]{forgetting}
Mariya Toneva, Alessandro Sordoni, Remi~Tachet des Combes, Adam Trischler,
  Yoshua Bengio, and Geoffrey~J. Gordon.
\newblock An empirical study of example forgetting during deep neural network
  learning.
\newblock In \emph{International Conference on Learning Representations}, 2019.

\bibitem[Touvron et~al.(2023)Touvron, Lavril, Izacard, Martinet, Lachaux,
  Lacroix, Rozi{\`e}re, Goyal, Hambro, Azhar, et~al.]{llama}
Hugo Touvron, Thibaut Lavril, Gautier Izacard, Xavier Martinet, Marie-Anne
  Lachaux, Timoth{\'e}e Lacroix, Baptiste Rozi{\`e}re, Naman Goyal, Eric
  Hambro, Faisal Azhar, et~al.
\newblock Llama: Open and efficient foundation language models.
\newblock \emph{arXiv preprint arXiv:2302.13971}, 2023.

\bibitem[Vicente et~al.(1994)Vicente, Savard, and J{\'u}dice]{bilevel_np_hard}
Luis Vicente, Gilles Savard, and Joaquim J{\'u}dice.
\newblock Descent approaches for quadratic bilevel programming.
\newblock \emph{Journal of Optimization theory and applications}, 81\penalty0
  (2):\penalty0 379--399, 1994.

\bibitem[Vicol et~al.(2021)Vicol, Metz, and
  Sohl-Dickstein]{bilevel_bias_variance}
Paul Vicol, Luke Metz, and Jascha Sohl-Dickstein.
\newblock Unbiased gradient estimation in unrolled computation graphs with
  persistent evolution strategies.
\newblock In \emph{International Conference on Machine Learning}, pp.\
  10553--10563. PMLR, 2021.

\bibitem[Vicol et~al.(2022)Vicol, Lorraine, Pedregosa, Duvenaud, and
  Grosse]{bilevel_implicit_bias}
Paul Vicol, Jonathan~P Lorraine, Fabian Pedregosa, David Duvenaud, and Roger~B
  Grosse.
\newblock On implicit bias in overparameterized bilevel optimization.
\newblock In \emph{International Conference on Machine Learning}. PMLR, 2022.

\bibitem[Wang et~al.(2022)Wang, Zhao, Peng, Zhu, Yang, Wang, Huang, Bilen,
  Wang, and You]{cafe}
Kai Wang, Bo~Zhao, Xiangyu Peng, Zheng Zhu, Shuo Yang, Shuo Wang, Guan Huang,
  Hakan Bilen, Xinchao Wang, and Yang You.
\newblock Cafe: Learning to condense dataset by aligning features.
\newblock In \emph{Proceedings of the IEEE/CVF Conference on Computer Vision
  and Pattern Recognition}, pp.\  12196--12205, 2022.

\bibitem[Wang et~al.(2021)Wang, Shivanna, Cheng, Jain, Lin, Hong, and
  Chi]{dcnv2}
Ruoxi Wang, Rakesh Shivanna, Derek Cheng, Sagar Jain, Dong Lin, Lichan Hong,
  and Ed~Chi.
\newblock Dcn v2: Improved deep \& cross network and practical lessons for
  web-scale learning to rank systems.
\newblock In \emph{Proceedings of the web conference 2021}, pp.\  1785--1797,
  2021.

\bibitem[Wang et~al.(2018)Wang, Zhu, Torralba, and Efros]{dd_orig}
Tongzhou Wang, Jun-Yan Zhu, Antonio Torralba, and Alexei~A Efros.
\newblock Dataset distillation.
\newblock \emph{arXiv preprint arXiv:1811.10959}, 2018.

\bibitem[Welling(2009)]{herding}
Max Welling.
\newblock Herding dynamical weights to learn.
\newblock In \emph{Proceedings of the 26th Annual International Conference on
  Machine Learning}, ICML '09, 2009.

\bibitem[Wiewel \& Yang(2021)Wiewel and Yang]{dd_continual_1}
Felix Wiewel and Bin Yang.
\newblock Condensed composite memory continual learning.
\newblock In \emph{2021 International Joint Conference on Neural Networks
  (IJCNN)}, pp.\  1--8. IEEE, 2021.

\bibitem[Wolf et~al.(2020)Wolf, Debut, Sanh, Chaumond, Delangue, Moi, Cistac,
  Rault, Louf, Funtowicz, Davison, Shleifer, von Platen, Ma, Jernite, Plu, Xu,
  Le~Scao, Gugger, Drame, Lhoest, and Rush]{huggingface}
T.~Wolf, L.~Debut, V.~Sanh, J.~Chaumond, C.~Delangue, A.~Moi, P.~Cistac,
  T.~Rault, R.~Louf, M.~Funtowicz, J.~Davison, S.~Shleifer, P.~von Platen,
  C.~Ma, Y.~Jernite, J.~Plu, C.~Xu, T.~Le~Scao, S.~Gugger, M.~Drame, Q.~Lhoest,
  and A.~Rush.
\newblock Transformers: State-of-the-art natural language processing.
\newblock In \emph{Proceedings of the 2020 Conference on Empirical Methods in
  Natural Language Processing: System Demonstrations}, pp.\  38--45, Online,
  October 2020. Association for Computational Linguistics.
\newblock \doi{10.18653/v1/2020.emnlp-demos.6}.

\bibitem[Wolpert \& Macready(1997)Wolpert and Macready]{no_free_lunch}
D.H. Wolpert and W.G. Macready.
\newblock No free lunch theorems for optimization.
\newblock \emph{IEEE Transactions on Evolutionary Computation}, 1\penalty0
  (1):\penalty0 67--82, 1997.
\newblock \doi{10.1109/4235.585893}.

\bibitem[Wu et~al.(2022)Wu, Raghavendra, Gupta, Acun, Ardalani, Maeng, Chang,
  Aga, Huang, Bai, et~al.]{data_increasing_recsys}
Carole-Jean Wu, Ramya Raghavendra, Udit Gupta, Bilge Acun, Newsha Ardalani,
  Kiwan Maeng, Gloria Chang, Fiona Aga, Jinshi Huang, Charles Bai, et~al.
\newblock Sustainable ai: Environmental implications, challenges and
  opportunities.
\newblock \emph{Proceedings of Machine Learning and Systems}, 4:\penalty0
  795--813, 2022.

\bibitem[Wu et~al.(2020)Wu, Sun, Zhang, Xie, and Cui]{gnn_recsys_survey}
Shiwen Wu, Fei Sun, Wentao Zhang, Xu~Xie, and Bin Cui.
\newblock Graph neural networks in recommender systems: a survey.
\newblock \emph{ACM Computing Surveys (CSUR)}, 2020.

\bibitem[Wu et~al.(2019)Wu, Tang, Zhu, Wang, Xie, and Tan]{gnn_recsys}
Shu Wu, Yuyuan Tang, Yanqiao Zhu, Liang Wang, Xing Xie, and Tieniu Tan.
\newblock Session-based recommendation with graph neural networks.
\newblock In \emph{Proceedings of the AAAI conference on artificial
  intelligence}, 2019.

\bibitem[Wu et~al.(2018)Wu, Ren, Liao, and Grosse.]{biased_bptt}
Yuhuai Wu, Mengye Ren, Renjie Liao, and Roger Grosse.
\newblock Understanding short-horizon bias in stochastic meta-optimization.
\newblock In \emph{International Conference on Learning Representations}, 2018.

\bibitem[Xiong et~al.(2022)Xiong, Wang, Cheng, Yu, and
  Hsieh]{federated_distill_4}
Yuanhao Xiong, Ruochen Wang, Minhao Cheng, Felix Yu, and Cho-Jui Hsieh.
\newblock Feddm: Iterative distribution matching for communication-efficient
  federated learning.
\newblock \emph{arXiv preprint arXiv:2207.09653}, 2022.

\bibitem[Xu \& Cohen(2018)Xu and Cohen]{stock}
Yumo Xu and Shay~B Cohen.
\newblock Stock movement prediction from tweets and historical prices.
\newblock In \emph{Proceedings of the 56th Annual Meeting of the Association
  for Computational Linguistics (Volume 1: Long Papers)}, pp.\  1970--1979,
  2018.

\bibitem[Zhao \& Bilen(2021)Zhao and Bilen]{zhao_dsa}
Bo~Zhao and Hakan Bilen.
\newblock Dataset condensation with differentiable siamese augmentation.
\newblock In \emph{International Conference on Machine Learning}, pp.\
  12674--12685. PMLR, 2021.

\bibitem[Zhao \& Bilen(2022)Zhao and Bilen]{gan_distillation}
Bo~Zhao and Hakan Bilen.
\newblock Synthesizing informative training samples with gan.
\newblock \emph{arXiv preprint arXiv:2204.07513}, 2022.

\bibitem[Zhao \& Bilen(2023)Zhao and Bilen]{dm}
Bo~Zhao and Hakan Bilen.
\newblock Dataset condensation with distribution matching.
\newblock In \emph{Proceedings of the IEEE/CVF Winter Conference on
  Applications of Computer Vision (WACV)}, 2023.

\bibitem[Zhao et~al.(2021)Zhao, Mopuri, and Bilen]{zhao_dc}
Bo~Zhao, Konda~Reddy Mopuri, and Hakan Bilen.
\newblock Dataset condensation with gradient matching.
\newblock In \emph{International Conference on Learning Representations}, 2021.

\bibitem[Zhou et~al.(2022{\natexlab{a}})Zhou, Pi, Zhang, Lin, Chen, and
  Zhang]{bilevel_coresets_bayesian}
Xiao Zhou, Renjie Pi, Weizhong Zhang, Yong Lin, Zonghao Chen, and Tong Zhang.
\newblock Probabilistic bilevel coreset selection.
\newblock In \emph{International Conference on Machine Learning}, pp.\
  27287--27302. PMLR, 2022{\natexlab{a}}.

\bibitem[Zhou et~al.(2020)Zhou, Pu, Ma, Li, and Wu]{federated_distill_2}
Yanlin Zhou, George Pu, Xiyao Ma, Xiaolin Li, and Dapeng Wu.
\newblock Distilled one-shot federated learning.
\newblock \emph{arXiv preprint arXiv:2009.07999}, 2020.

\bibitem[Zhou et~al.(2022{\natexlab{b}})Zhou, Nezhadarya, and Ba]{frepo}
Yongchao Zhou, Ehsan Nezhadarya, and Jimmy Ba.
\newblock Dataset distillation using neural feature regression.
\newblock In \emph{Advances in Neural Information Processing Systems},
  2022{\natexlab{b}}.

\end{thebibliography}
\bibliographystyle{tmlr}

\newpage

\appendix

\section{Notation} \label{appendix:details}

\def\arraystretch{1.5}
\begin{table*}[!htbp]
\begin{center}
\begin{tabular}{r@{\hskip 0.4in} l}
    \multicolumn{2}{c}{\textbf{Dataset related}} \\[0.1in]
    
    $\dataset \triangleq \{ (x_i \in \mathcal{X}, y_i \in \mathcal{Y}) \}_{i=1}^{|\dataset|}$ & The target dataset to be distilled \\
    $\mathcal{X}$ & Data domain \\
    $\mathcal{Y}$ & Predictand domain \\
    $\mathcal{C}$ & Set of unique classes in $\mathcal{Y}$ \\
    $\dataset^c \triangleq \{ (x_i, y_i) ~|~ y_i = c\}_{i=1}^{|\dataset|}$ & Portion of \dataset with class $c$ \\
    $\mathbf{X} \triangleq [x_i]_{i=1}^{|\dataset|}$ & Matrix of all features in \dataset \\
    $\mathbf{Y} \triangleq [y_i]_{i=1}^{|\dataset|}$ & Matrix of all predictands in \dataset \\
    $n$ & Size of data summary \\
    $\distill \triangleq \{ (\Tilde{x}_i, \Tilde{y}_i) \}_{i=1}^{n}$ & Data summary \\
    $\distill^c \triangleq \{ (\Tilde{x}_i, \Tilde{y}_i) ~|~ \Tilde{y}_i = c\}_{i=1}^{n}$ & Portion of \distill with class $c$ \\
    $\mathbf{X}_{\mathsf{syn}} \triangleq [\Tilde{x}_i]_{i=1}^{n}$ & Matrix of all features in \distill \\
    $\mathbf{Y}_{\mathsf{syn}} \triangleq [\Tilde{y}_i]_{i=1}^{n}$ & Matrix of all predictands in \distill

    \\[0.1in] \multicolumn{2}{c}{\textbf{Learning related}} \\[0.1in]
    
    $\Phi_{\theta} : \mathcal{X} \mapsto \mathcal{Y}$ & Learning algorithm parameterized by $\theta$ \\
    $l : \mathcal{Y} \times \mathcal{Y} \mapsto \mathbb{R}$ & Twice-differentiable cost function \\
    $\mathcal{L}_{\dataset}(\theta) \triangleq \mathbb{E}_{(x, y) \sim \dataset}[l(\Phi_{\theta}(x), y)]$ & Expected loss of $\Phi$ on \dataset \\
    $\mathcal{L}_{\distill}(\theta) \triangleq \mathbb{E}_{(x, y) \sim \distill}[l(\Phi_{\theta}(x), y)]$ & Expected loss of $\Phi$ on \distill 
    
    \\[0.1in] \multicolumn{2}{c}{\textbf{General}} \\[0.1in]

    $\dim(\mathcal{A})$ & Size of basis of $\mathcal{A}$ \\
    $|\mathcal{A}|$ & Number of elements in $\mathcal{A}$ \\
    $\operatorname{sup}$ & Supremum \\
    $\underset{\theta}{\operatorname{arg} \operatorname{min}} ~~ f(\theta)$ & Optimum value of $\theta$ which minimizes $f(\theta)$ \\
    $\expectation{x}{f(x)} \triangleq \sum_x p(x) \cdot f(x)$ & Expected value of $f(x)$ when domain of $x$ is discrete
\end{tabular}
\end{center}
\end{table*}

% \vspace{0.25cm}
% \newpage

% \section{Collated Results} \label{appendix:results}

% \input{results.tex}

\end{document}